\documentclass[10pt,twocolumn,letterpaper]{article}
\usepackage{tikz}
\usepackage{everypage}
\AddThispageHook{%
  \begin{tikzpicture}[remember picture,overlay]
    \node[above left, inner sep=0pt] at
         ([xshift=-1.35cm,yshift=-1.5cm]current page.north east) % 微调偏移
         {\includegraphics[height=1cm]{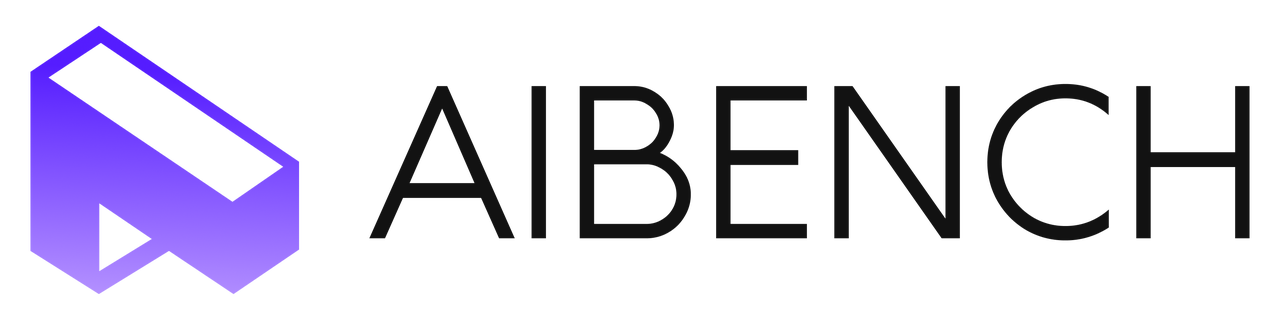}};             % Logo 文件名
  \end{tikzpicture}%
}

\usepackage[pagenumbers]{cvpr} % To force page numbers, e.g. for an arXiv version
\usepackage{textpos}
\definecolor{cvprblue}{rgb}{0.21,0.49,0.74}
\usepackage[pagebackref,breaklinks,colorlinks,allcolors=cvprblue]{hyperref}

\usepackage{fontawesome}
\usepackage{multirow}
\usepackage{arydshln}
\usepackage{comment}

\def\confYear{2026}
\usepackage{xcolor}

\definecolor{red}{RGB}{255, 85, 85}  
\definecolor{blue}{RGB}{135, 206, 250} 
\definecolor{green}{RGB}{152, 251, 152}

\definecolor{lightgray}{gray}{0.9}
\newcommand{\CLA}[1]{{\color[HTML]{4472c4} \textbf{#1}}}
\newcommand{\CLB}[1]{{\color[HTML]{E76254} \textbf{#1}}}
\newcommand{\CLC}[1]{{\color[HTML]{87ADE0} \textbf{#1}}}
\newcommand{\CLD}[1]{{\color[HTML]{FFA391} \textbf{#1}}}

\newcommand{\GroupA}[1]{{\color[HTML]{92ca2c} \textbf{#1}}}
\newcommand{\GroupB}[1]{{\color[HTML]{3c7daa} \textbf{#1}}}
\newcommand{\GroupC}[1]{{\color[HTML]{e64f04} \textbf{#1}}}
\newcommand{\GroupD}[1]{{\color[HTML]{f2c400} \textbf{#1}}}
\newcommand{\GroupE}[1]{{\color[HTML]{a757a7} \textbf{#1}}}

\title{Embodied Image Compression}

\author{Chunyi Li$^{1,2,3}$$^{*}$, Rui Qing$^{1}$$^{*}$, Jianbo Zhang$^{1}$, Yuan Tian$^{2}$, Xiangyang Zhu$^{2}$,\\ Zicheng Zhang$^{1,2}$, Xiaohong Liu$^{1}$, Weisi Lin$^{3}$, Guangtao Zhai$^{1,2}$\\
Shanghai Jiao Tong University$^{1}$, Shanghai AI Lab$^{2}$, Nanyang Technological University$^{3}$\\
}
\begin{document}
% \maketitle

\twocolumn[{%
\renewcommand\twocolumn[1][]{#1}%
\maketitle
\begin{center}
    \centering
    \vspace{-2.4em}
    \includegraphics[width=1\linewidth]{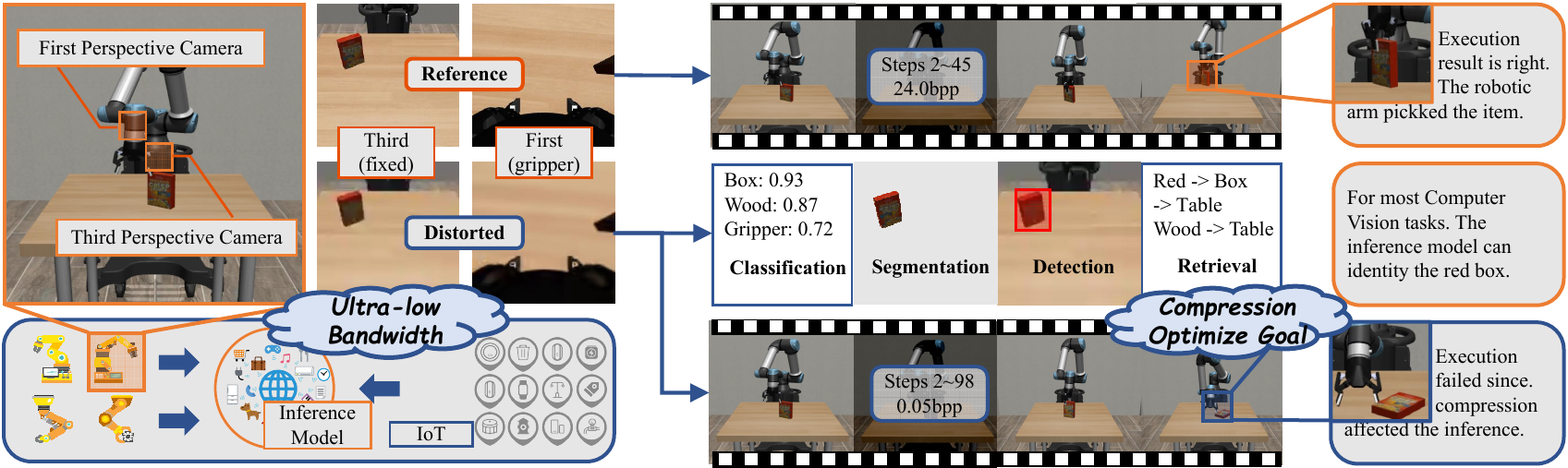}
    \vspace{-20pt}
    \captionof{figure}{In Embodied AI inference, images need to be compressed to meet the edge-cloud bandwidth limitations. However, due to the differences between \CLB{Machine Visual System (MVS)} and \CLA{Robotic Visual System (RVS)}, although existing compression metrics can maintain fidelity on Computer Vision (CV) tasks, they can lead to serious errors for Embodied AI manipulation, vice versa.} 
    \label{fig:spotlight}
    
\end{center}%
}]

\newcommand\blfootnote[1]{%
\begingroup
\renewcommand\thefootnote{}\footnote{#1}%
\addtocounter{footnote}{-1}%
\endgroup
}

\begin{abstract}

Image Compression for Machines (ICM) has emerged as a pivotal research direction in the field of visual data compression. However, with the rapid evolution of machine intelligence, the target of compression has shifted from task-specific virtual models to Embodied agents operating in real-world environments. To address the communication constraints of Embodied AI in multi-agent systems and ensure real-time task execution, this paper introduces, for the first time, the scientific problem of Embodied Image Compression. We establish a standardized benchmark, EmbodiedComp, to facilitate systematic evaluation under ultra-low bitrate conditions in a closed-loop setting. Through extensive empirical studies in both simulated and real-world settings, we demonstrate that existing Vision-Language-Action models (VLAs) fail to reliably perform even simple manipulation tasks when compressed below the Embodied bitrate threshold. We anticipate that EmbodiedComp will catalyze the development of domain-specific compression tailored for Embodied agents
, thereby accelerating the Embodied AI deployment in the Real-world.

%\CLB{We will open-source all data for non-commercial use.}

% \CLB{Supplementary are attached at the bottom.}

\end{abstract}

\section{Introduction}

The target of visual signal compression has shifted from human perception to machine consumption. According to the Cisco \cite{review:cisco} white paper, the number of Machine-to-Machine (M2M) connections first exceeded that of Machine-to-Human (M2H) in 2023, reaching 147 billion. Consequently, since 2020, standards bodies such as ITU-T \cite{realte:icm-mpeg} have introduced Image/Video Compression for Machine (ICM/VCM) \cite{add:li1,add:li2,add:gao1,add:pku1,add:pku2,add:msra1,add:msra2,add:tian1,add:tian2,add:tian3,add:zhang1}. These standards optimize visual signal representation for downstream machine-task performance \cite{relate:icm-survey} rather than for human subjective quality. Then, with the rise of Embodied AI, the `Machine' in ICM has evolved from generic algorithms (e.g., segmentation, detection) to Real-world robotic systems.

\begin{table*}[t]
\centering
    \caption{Comparison with EmbodiedComp and existing Image Compression for Machine (ICM) frameworks. Unlike directly using CV datasets, all our data is \CLA{self-collected} through simulation rendering/real-world experiments. In this \CLA{closed-loop paradigm}, not only does compression affect inference results, but the actions of Embodied AI also influence subsequent compression iterations.}
    \label{tab:relate}
    \vspace{-8pt}
    \renewcommand\arraystretch{1.4}
    \belowrulesep=0pt\aboverulesep=0pt
    \resizebox{\linewidth}{!}{
\begin{tabular}{llccccc}
\toprule
Name         & Source   & Dataset                                  & Task                             & Index                & Serve          & Evaluation   \\ \midrule
UG-ICM \cite{relate:icm-UG-ICM}      & AAAI 25  & ILSVRC 12; PASCAL VOC; COCO 17; Kodak    & IQA; CLS; DET; SEG               & Acc; mAP; mIOU; PSNR & Human, Machine & open-loop    \\
DICM \cite{relate:icm-DICM}        & TBC 25   & ImageNet; PASCAL VOC; COCO 17; COCO 14   & CLS; DET; SEG                    & Acc; AP; mAP         & Machine        & open-loop    \\
NPP \cite{relate:icm-NPP}         & TCSVT 24 & ImageNet; COCO 17;                       & CLS; DET                         & Acc; mAP             & Machine        & open-loop    \\
RDCC \cite{relate:icm-RDCC}        & ECCV 24  & ImageNet-1k; COCO 17; CityScapes         & CLS; DET; SEG                    & Acc; AP; mIOU        & Machine        & open-loop    \\
ICMH-Net \cite{relate:icm-ICMH-Net}    & ACMMM 23 & ILSVRC 12; PASCAL VOC; COCO 14           & CLS; DET; SEG                    & Acc; AP; mIOU        & Machine        & open-loop    \\
GISwin-Block \cite{relate:icm-GISwin-Block} & ICCV 23  & COCO 17~                                 & SEG; RET; IQA                         & AP; PSNR                   & Human, Machine        & open-loop    \\
TransTIC \cite{relate:icm-TransTIC}    & ICCV 23  & ImageNet; COCO 17; COCO 14               & CLS; DET; SEG                    & Acc; AP; mAP         & Machine        & open-loop    \\
OmniICM \cite{relate:icm-OmniICM}     & ECCV 22  & PASCAL VOC; COCO 17; COCO 14; CityScapes & CLS; DET; SEG; RET               & AP; mAP; mIOU        & Machine        & open-loop    \\ \cdashline{1-7}
EmbodiedComp & ours     & Self-collected in Simulation and Real-World                          & Pi0.5, Pi0, OpenVLA Manipualtion & SR, Step             & Robotics       & cloesd-loop \\ \bottomrule
\end{tabular}}
\vspace{-2mm}
\end{table*}

Compared with general-purpose Computer Vision (CV), dedicated compression for Embodied AI is imperative for three reasons. 
($\romannumeral1$) Embodied AI is more communication-dependent than conventional vision systems.  In traditional setups the imager and the compute unit are co-located, allowing direct on-device inference; in Embodied industrial scenarios, however, the robotic arm and the camera are physically separated, forcing visual data to be transmitted before processing. 
($\romannumeral2$) Embodied platforms operate under markedly narrower communication budgets.  Laboratory demonstrations often assume a single, dedicated link, yet real-world deployments place multiple agents within a shared Internet-of-Things (IoT) backbone whose bandwidth is already scarce.  Empirically, images must be compressed to 0.1\% of their original size to meet channel constraints.
($\romannumeral3$) The optimization goal during compression diverges from that of standard CV tasks.  Preservation fidelity above 95\% for segmentation or detection benchmarks does not guarantee that an Embodied agent can still execute user commands from the compressed stream, and vice-versa.
As illustrated in Figure \ref{fig:spotlight}, Embodied Image Compression constitutes a problem class distinct from previous ICM: it targets an emerging Robotic Visual System (RVS) at drastically lower bit-rates, rather than serving the well-explored Human and Machine Visual System (HVS/MVS).

Therefore, recognizing Embodied systems routinely operate under severe bandwidth constraints in Real-world deployments, we introduce the task of Embodied Image Compression. Our contributions can be summarized as follows:

\begin{itemize} 
\item EmbodiedComp benchmark: We release the first dataset tailored to Embodied manipulation, comprising 100 standardized test sequences rendered with varied object layouts, backgrounds, and environmental states. Using 2,000 manipulation trajectories, we train three Vision–Language–Action models (VLAs) that all achieve optimal when supplied with uncompressed imagery.
\item Theoretical modeling: We derive the RVS-bitrate relationship for Embodied perception. In contrast to HVS and MVS, the RVS exhibits graceful degradation under light compression but undergoes an abrupt performance collapse once the bitrate falls below a critical threshold.
\item Empirical evaluation: We validate 10 advanced image codecs on Embodied manipulation tasks. Sim2real experiments demonstrate that none of the three VLAs maintains operational status on our EmbodiedComp when fed compressed images, underscoring the urgent need for codecs explicitly designed for Embodied AI.
\end{itemize}

\section{Related Works}

\subsection{Embodied AI Manipulation}

Embodied intelligence has advanced rapidly and spawned a wide range of applications, yet public demonstrations still rely on either ($\romannumeral1$) fully co-located sensing-and-compute modules that avoid wireless communication, or ($\romannumeral2$) idealized links with abundant bandwidth—conditions that exist only in laboratories. Practical deployment inevitably requires efficient, Real-time agentic communication: edge devices acquire the imagery and the cloud performs inference, making compression a mandatory step in the loop.

Embodied tasks fall into two broad categories: manipulation and navigation, where the Edge2Cloud paradigm above is applied predominantly only to manipulation.  First, navigation models can be deployed on-board, whereas manipulation typically demands multi-view cameras, so wireless transmission becomes unavoidable. Second, navigation is less mature—humanoid/quadruped platforms cope only with simple obstacle avoidance, so compression deployment is not yet ready. Manipulation, by contrast, has produced comparatively robust models such as CogACT \cite{vla:cogact}, DreamVLA \cite{vla:dreamvla}, Octo \cite{vla:octo}, and Pi0 \cite{vla:pi0}, which generalize within fixed scenes and thus constitute a realistic test-bed for compression research.
These VLAs uniformly accept images as input and output 7/8-DoF end-effector poses. When video streams are supplied, latency constraints preclude frame-wise pose estimation; instead, a small set of key frames is analyzed offline. Consequently, the present study focuses on manipulation tasks, employs still-image rather than video codecs to compress the visual signal, and validates the pipeline through downstream VLA inference.

\begin{figure*}
\centering
\includegraphics[width = \linewidth]{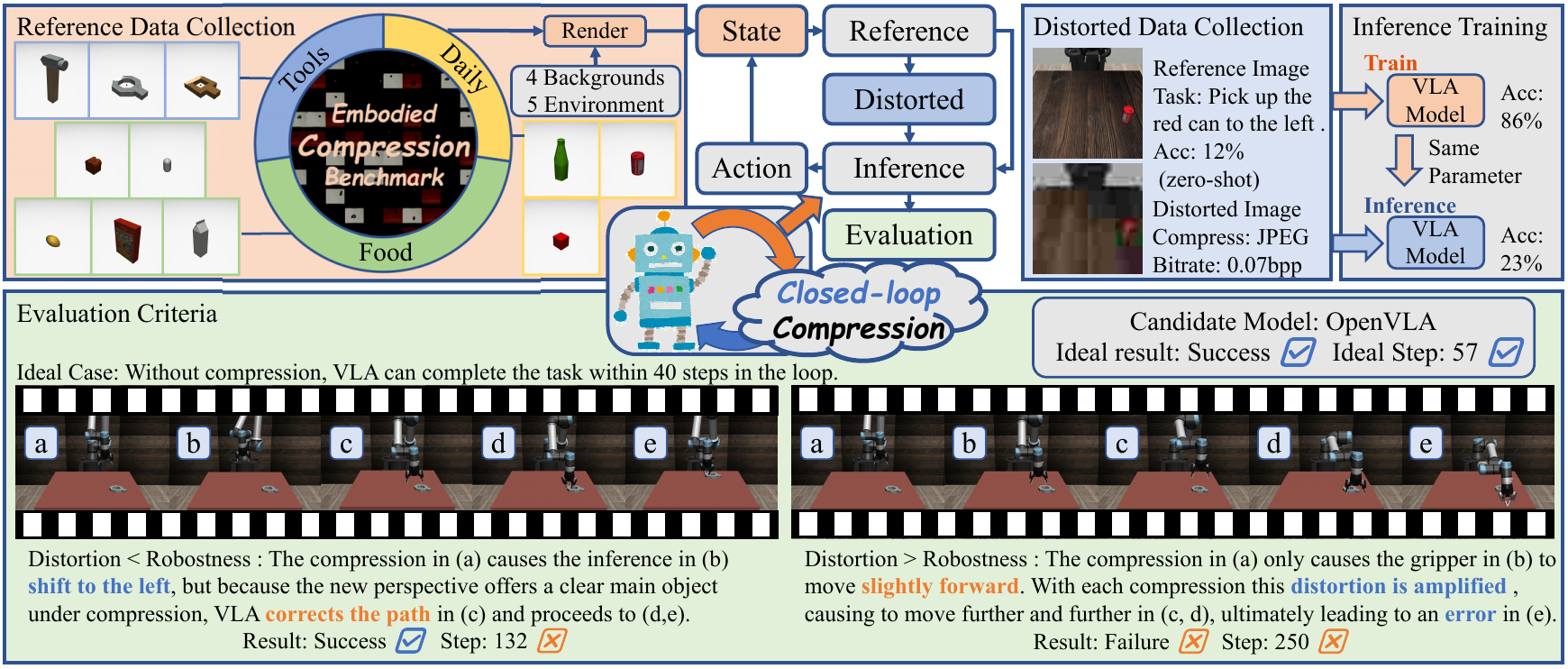}
%\vspace{-2mm}
\caption{Overview of EmbodiedComp benchmark. To align with Real-world applications of Embodied AI, we deploy the compression algorithm within the Embodied Inference pipeline for the first time, enabling closed-loop validation. Since compression distortion accumulates in each loop-iterations, evaluation metrics include both Success Rate (SR) and the Step for iterations to represent efficiency.}
\vspace{-3mm}
\label{fig:dataset}
\end{figure*}

\subsection{Image Compression for Machine}

Traditional ICM frameworks in Table \ref{tab:relate} are designed for the HVS/MVS. All of them operate in open-loop: after compression, the downstream task is executed immediately thereafter. In contrast, RVS follows a closed-loop, multi-step cycle: `Sample→Compression→Action→State Change→Resample'. Moreover, existing datasets are evaluated at high bitrates using standard CV tasks and simply reuse legacy corpora such as COCO \cite{relate:coco} or ImageNet \cite{relate:imagenet}. Although recent works has introduced datasets tailored for Vision-Language Model (VLM) \cite{relate:icm-mllm1,relate:icm-mllm2,relate:icm-mllm3,relate:icm-mllm4,relate:icm-mllm5,relate:icm-mllm6} or at ultra-low bitrates \cite{realte:ultra1,relate:ultra2,relate:ultra3,relate:ultra4} with high signal fidelity \cite{add:hu1,add:hu2}, no dataset to date employs a VLA as the final receiver. Consequently, compression for Embodied AI must be rebuilt from the ground up, starting with the data itself.

\section{Benchmark Construction}

As illustrated in Figure \ref{fig:dataset}, the EmbodiedComp closed-loop comprises four sequential modules:
 ($\romannumeral1$) Rendering and sampling a reference image in the simulation, ($\romannumeral2$) Feeding reference image into the codec for a certain bitrate distortion, ($\romannumeral3$) Forwarding distorted image to the VLA; this inference result will be executed in the simulation, thereby changing the state of the gripper itself and the external environment, thus returning to ($\romannumeral1$) sample a new image. When the success condition is met or the upper iteration limit is exceeded, the loop will be exited for ($\romannumeral4$) Evaluation. This section will analyze the above modules separately.

\subsection{Reference Data Collection}
\label{sec:ref}

The EmbodiedComp simulation stack is built on Robosuite 1.5.1 with MuJoCo 3.3.4 as the physics backend. The experimental scene is instantiated through a custom LiftTest class that contains a textured background, a rigid tabletop, and manipulable objects on it. A codec operating inside this loop first samples the current reference image rendered by the simulator, then compresses it to the target bitrate. Unlike conventional ICM benchmarks—which compress a fixed corpus of reference images—EmbodiedComp performs dynamic rendering: the image to be compressed does not exist until the previous VLA action modifies the environment.
Consequently, compression and VLA inference are mutually causal: ($\romannumeral1$) compression distortion degrades VLA policy accuracy, and ($\romannumeral2$) the pose produced by the VLA determines the next scene configuration and hence the next image that enters the compressor. The simulation scene includes:
\begin{itemize}
    \item Main object: Common (Bottle, Can, Cube), Food (Bread, Capsule, Cereal, Lemon, Milk), and Tools (Hammer, Nut round, Nut square);
    \item Table: Black, Ceramic, Cherry, Wood dark, Wood light;
    \item Background: Daily, Dark, Light, Wall.
\end{itemize}
EmbodiedComp is split into a train and a test split with disjoint simulation scenes.
The train split is used only for VLA fine-tuning to ensure high success rates on downstream uncompressed imagery; the test split is reserved for codec evaluation, i.e., to measure whether the VLA can still execute the correct action after compression. Both splits contain natural-language commands that refer exclusively to a single main object.
To isolate the effect of compression distortion from policy limitations, we restrict the command space to three primitive actions: pick, push, and press, since current VLAs cannot perform difficult flexible movements even when uncompressed.
Train split consists 2,000 static expert trajectories collected in the aforementioned Robosuite environment. A human operator successively issued a command, performed the corresponding motion, and the full state-action sequence was logged.
Test split includes 100 fully interactive scenes\footnote{Here 100 does not mean we only compressed 100 images, but rather that we validated 100 sences. EmbodiedComp accumulate 10,000$\sim$50,000 compressed frames, which depends on how quickly the VLA succeeds or fails in each scene, providing a statistically reliable estimation of codec.} that serve as the initial state for the closed-loop evaluation protocol. Each scene is repeatedly rendered in a fixed Third-person camera and a First-person perspective on gripper, compressed by certian codecs test, and acted upon by the fine-tuned VLA until the task succeeds or the iteration step budget is exhausted.
% 这里的100不代表我们只压缩了100张图像，而是验证了100个sence。总体调用次数取决于压缩算法的性能，在5000到25000之间。
% EmbodiedComp的仿真实验采用 Robosuite(1.5.1) 搭建，底层物理引擎为 MuJoCo(3.3.4)，环境基于自定义的 LiftTest 类，场景由背景、桌面、和放置在桌面上的可抓取物体组成。压缩算法会在这套框架下，sample原始的reference image用于进一步处理。可以看出，它和传统ICM的最大区别在于实时渲染，不是根据一组reference image来压缩，而是在环境中sample。因此，VLA与压缩是相互关联的（而非单向因果），不仅压缩失真会干扰VLA的表现，VLA的推理同样会决定下一张被压缩的图像。具体来说，场景信息包括：
% EmbodiedComp分为训练与测试集。训练集用于微调VLA，保证其在测试集上具有较高的成功率；测试集用来加压缩算法，看压了之后的VLA还能不能执行对。EmbodiedComp环境基于自定义的 LiftTest 类。场景由一个桌面和一个放置在桌面的可抓取物体组成
% 训练集是静态的开环数据，包含2,000真机操作轨迹。
% 具体来说，我们考虑了11种主体物，5种桌面，4种背景。

\begin{figure}
\centering
\includegraphics[width = \linewidth]{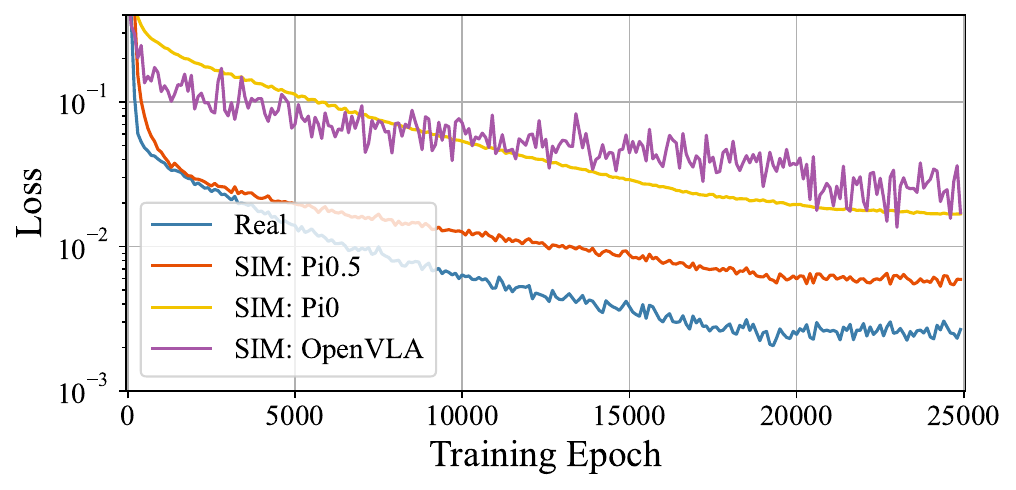}
\vspace{-7mm}
\caption{Training loss curve for Pi0.5, Pi0-Fast, and OpenVLA in simulation, and the Real-world model. All curves converge after 20,000 epochs, ensuring the performance before compression.}
\vspace{-3mm}
\label{fig:train}
\end{figure}

\subsection{Distorted Data Collection}
\label{sec:dis}

After acquiring images from First/Third-person perspectives, compression requires considering the Real-world situation of Embodied AI, and thus setting the highest possible bitrate-per-pixel (bpp) while ensuring Real-time communication. First, based on Shannon formula, the bitrate-per-second (bps) can be obtained as:
% 作为M2M的连接，按照现有的NB-IoT协议，Embodied AI通常在180kHz下通信。

\begin{equation}
    {\rm bps}=\frac{B}{||A||}{\rm log_2}(1+{\rm SNR}), 
\end{equation}
where $B$ denotes the bandwidth, as an M2M connection, Embodied AI communicates at 180kHz according to current NB-IoT \cite{review:NB-IoT} protocol. $A$ represents all agents in the network, typically in high Signal-to-Noise Ratio (SNR) environments such as indoors. Thus bpp can be estimated:

\begin{equation}
    {\rm bpp}=T \cdot \frac{S \cdot \rm bps} {2} \cdot \frac{1} {h \times w}
\end{equation}
where $T$ refers to transmission time. Considering current speed of VLA, it is set to 100ms for Real-time inference. $S$ denotes spectrum efficiency, which approximates 1 since current channel coding is already close to the Shannon limit at high SNR. $h,w$ stand for image height and width, which are $256\times256$ according the VLA input size. Then, to obtain the decodec image $\overline{I}$ compression will operate as:

\begin{equation}
    \overline{I} = {\mathcal C}_{q}({\mathcal D}_{r} (I)),\quad s.t. \overline{\rm  bpp}\leq{\rm  bpp}
\end{equation}
where ${\mathcal C}(\cdot),{\mathcal D}(\cdot)$ indicate Codec and Downsampling for the reference image $I$. This link will firstly try to adjust the quality $q$. If the $\overline{\rm  bpp}$ under the lowest $q$ still beyond the target bpp, then reduce the resolution $r$ until it meets requirements.
According to the above formula, for the indoor SNR=25dB and $||A||=10$ IoT devices, the bpp may reach 0.114; while for the lowest antenna allowable limit SNR=15dB and the upper IoT gateway limit $||A||=50$, the bpp will be 0.013. Therefore, all compressions in the EmbodiedCodec benchmark will operate within this range, where the target bpp is set as [0.015, 0.03, 0.06, 0.1].

\begin{figure}
\centering
\begin{minipage}[]{0.32\linewidth}
  \centering
  \centerline{\includegraphics[width = \textwidth]{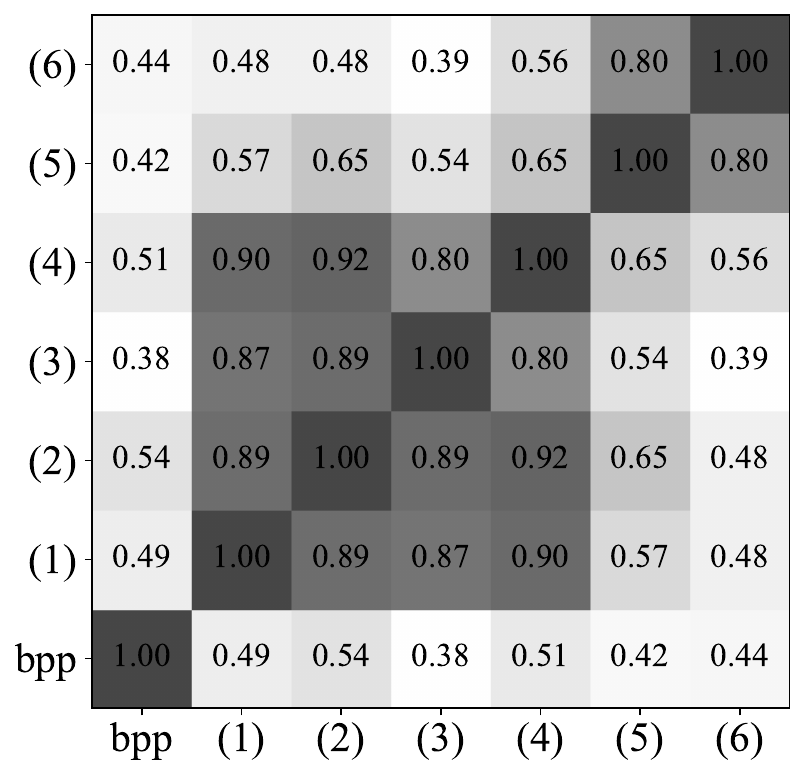}}
\end{minipage}
\begin{minipage}[]{0.32\linewidth}
  \centering
  \centerline{\includegraphics[width = \textwidth]{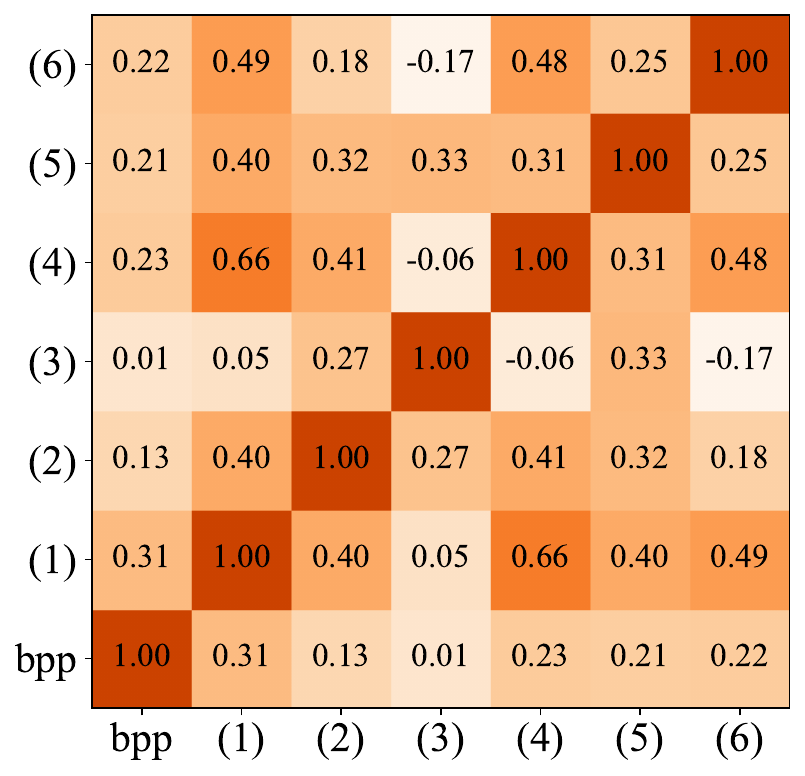}}
\end{minipage}
\begin{minipage}[]{0.32\linewidth}
  \centering
  \centerline{\includegraphics[width = \textwidth]{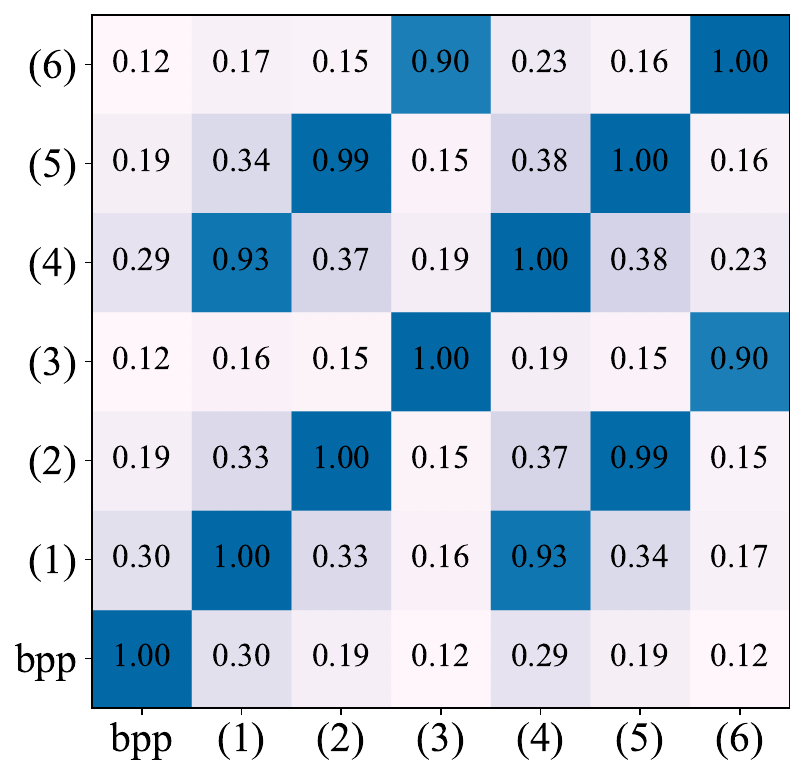}}
\end{minipage}

\begin{minipage}[]{0.32\linewidth}
  \centering
  \centerline{\includegraphics[width = \textwidth]{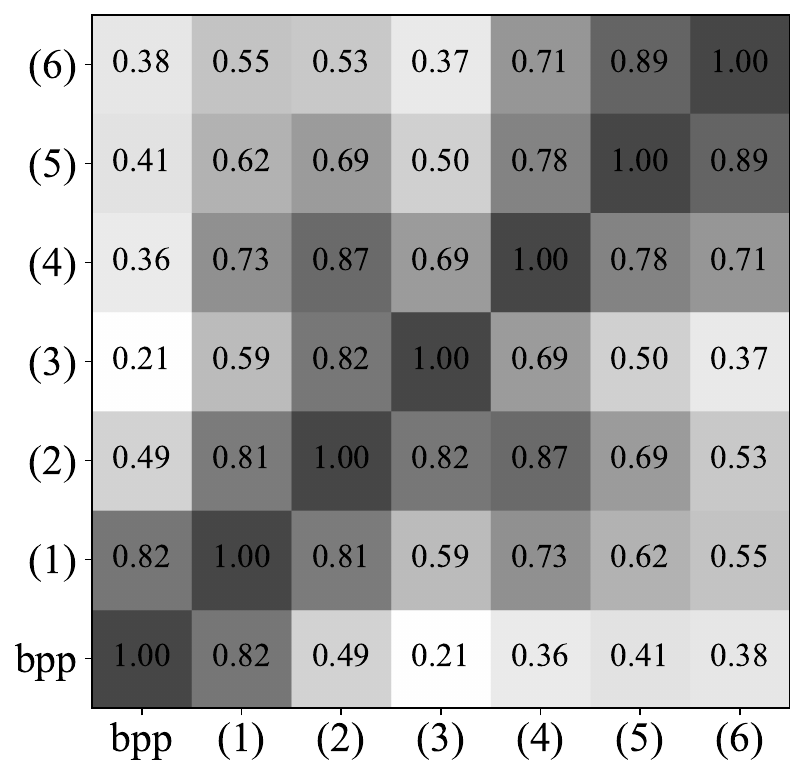}}
  \footnotesize \centerline{\textbf{HVS (0.53)}}\medskip
\end{minipage}
\vspace{-2mm}
\begin{minipage}[]{0.32\linewidth}
  \centering
  \centerline{\includegraphics[width = \textwidth]{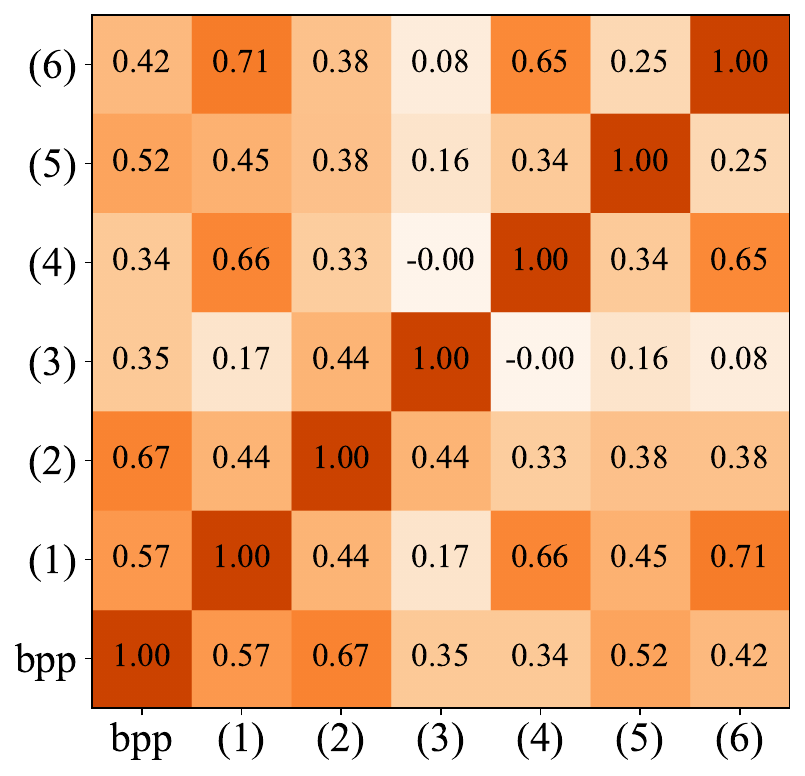}}
  \footnotesize \centerline{\CLB{MVS (0.41)}}\medskip
\end{minipage}
\vspace{-2mm}
\begin{minipage}[]{0.32\linewidth}
  \centering
  \centerline{\includegraphics[width = \textwidth]{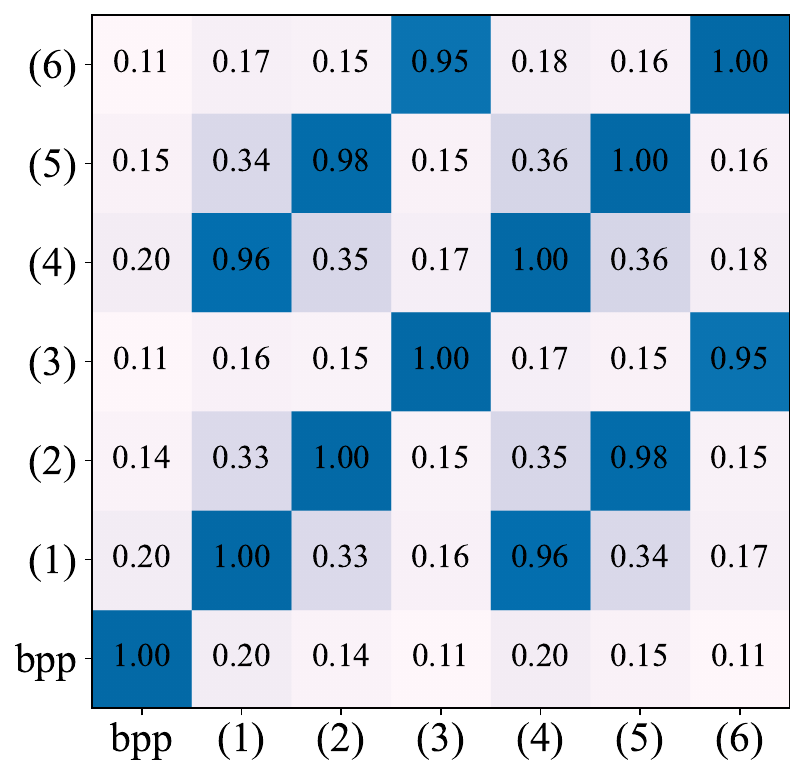}}
  \footnotesize \centerline{\CLA{RVS (0.15)}}\medskip
\end{minipage}
\vspace{-2mm}

\vspace{2mm}
\caption{SRCC (above) and PLCC (below) correlation matrix for HVS, MVS, and RVS-oriented indicator. RVS and bpp is most weakly correlated, reveals its difference against HVS/MVS.}
\vspace{-4mm}
\label{fig:corr}
\end{figure}

\subsection{Inference Model Training}
\label{sec:train}

Because zero-shot VLAs rarely succeed in unseen environments, EmbodiedComp first fine-tunes each candidate model on the training split until it is fully adapted to our scene layout and command distribution.  The resulting weights are frozen and reused at test time, guaranteeing that the VLA can already accomplish every task when supplied with the uncompressed reference image.  Any subsequent execution failure can therefore be attributed unambiguously to compression distortion rather than to a fundamental policy deficiency.
We select three representative VLAs as downstream validators:
Pi-0.5 \cite{vla:pi05} (best accuracy),
OpenVLA \cite{vla:openvla} (highest popularity), and
Pi0-Fast \cite{vla:pi0_fast} (fastest inference latency).
For Real-world experiments, whose visual complexity exceeds the simulation domain, we deploy only the strongest Pi-0.5.
Figure \ref{fig:train} shows training loss plateaus after 18,000-20,000 epochs for all three models, confirming convergence. The gap between uncompressed and compressed success rates thus serves as a clean proxy for the detrimental impact of compression on the RVS.

\begin{figure*}

\newcommand{\subfigoverlap}[2]{%
  \makebox[0.16\linewidth][l]{%
  \includegraphics[width=0.19\linewidth]{#1}%
    \hspace{-#2}%
  }%
}

\noindent
\subfigoverlap{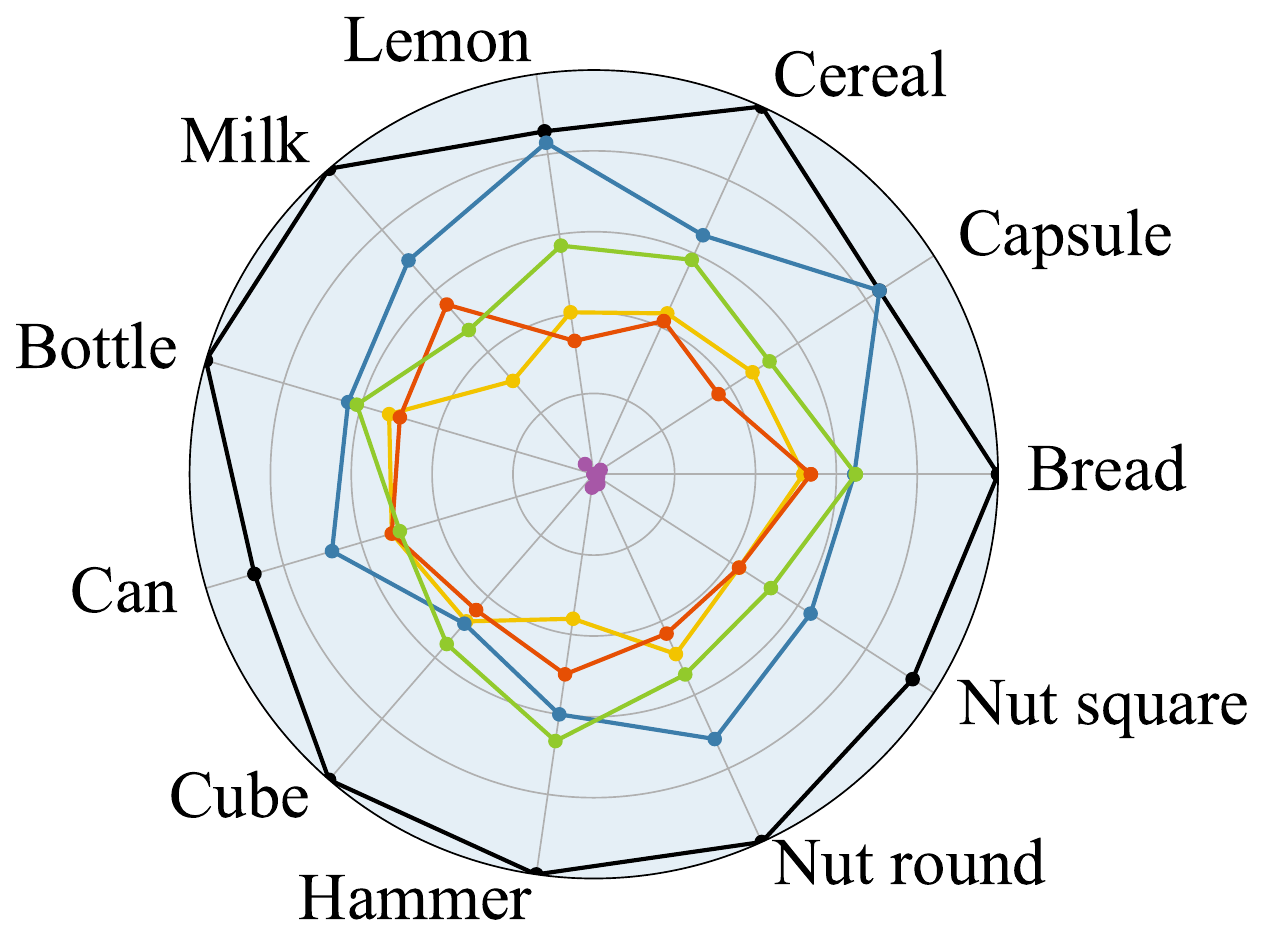}{5mm}%
\subfigoverlap{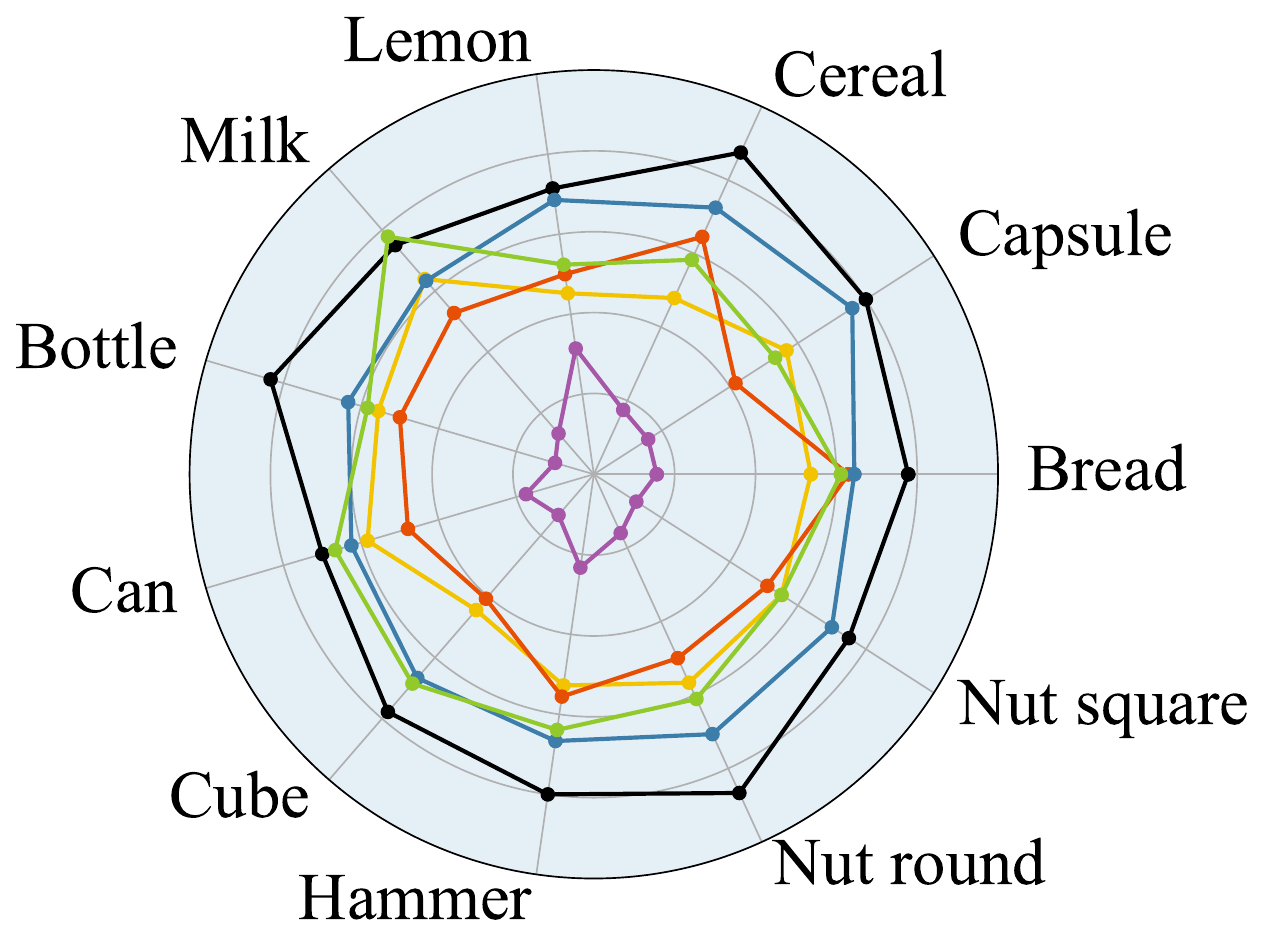}{5mm}%
\subfigoverlap{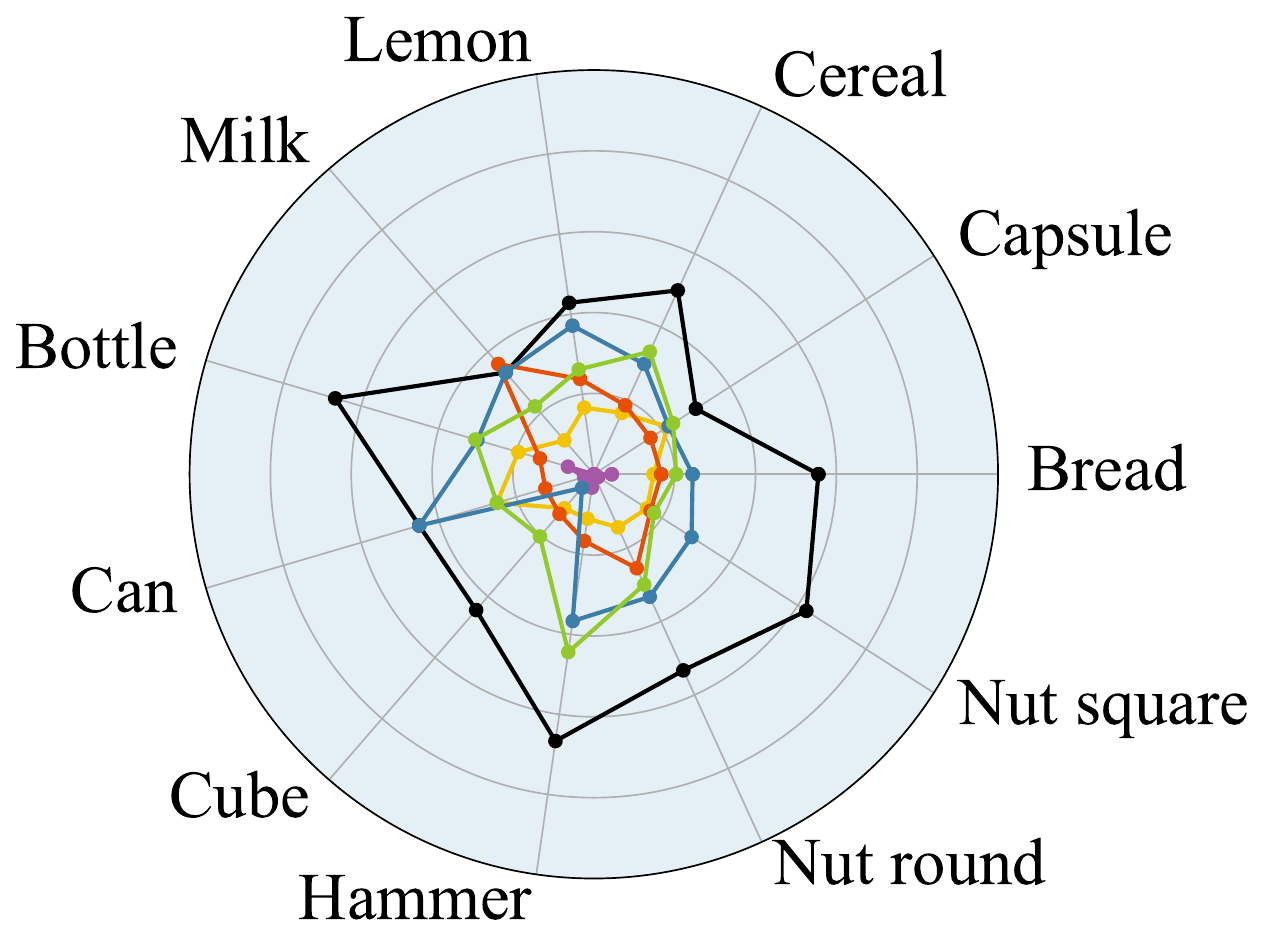}{5mm}
\subfigoverlap{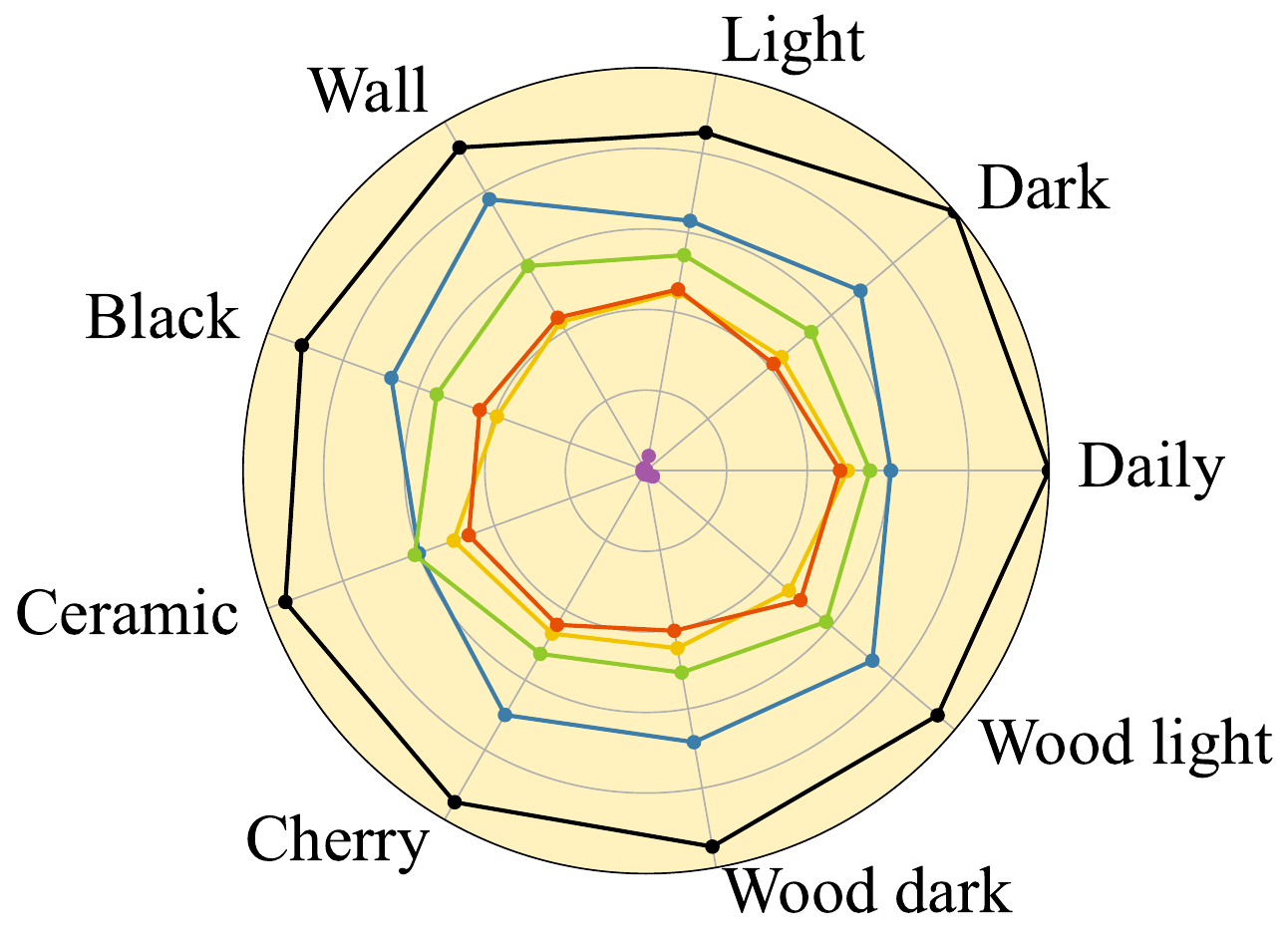}{5mm}%
\subfigoverlap{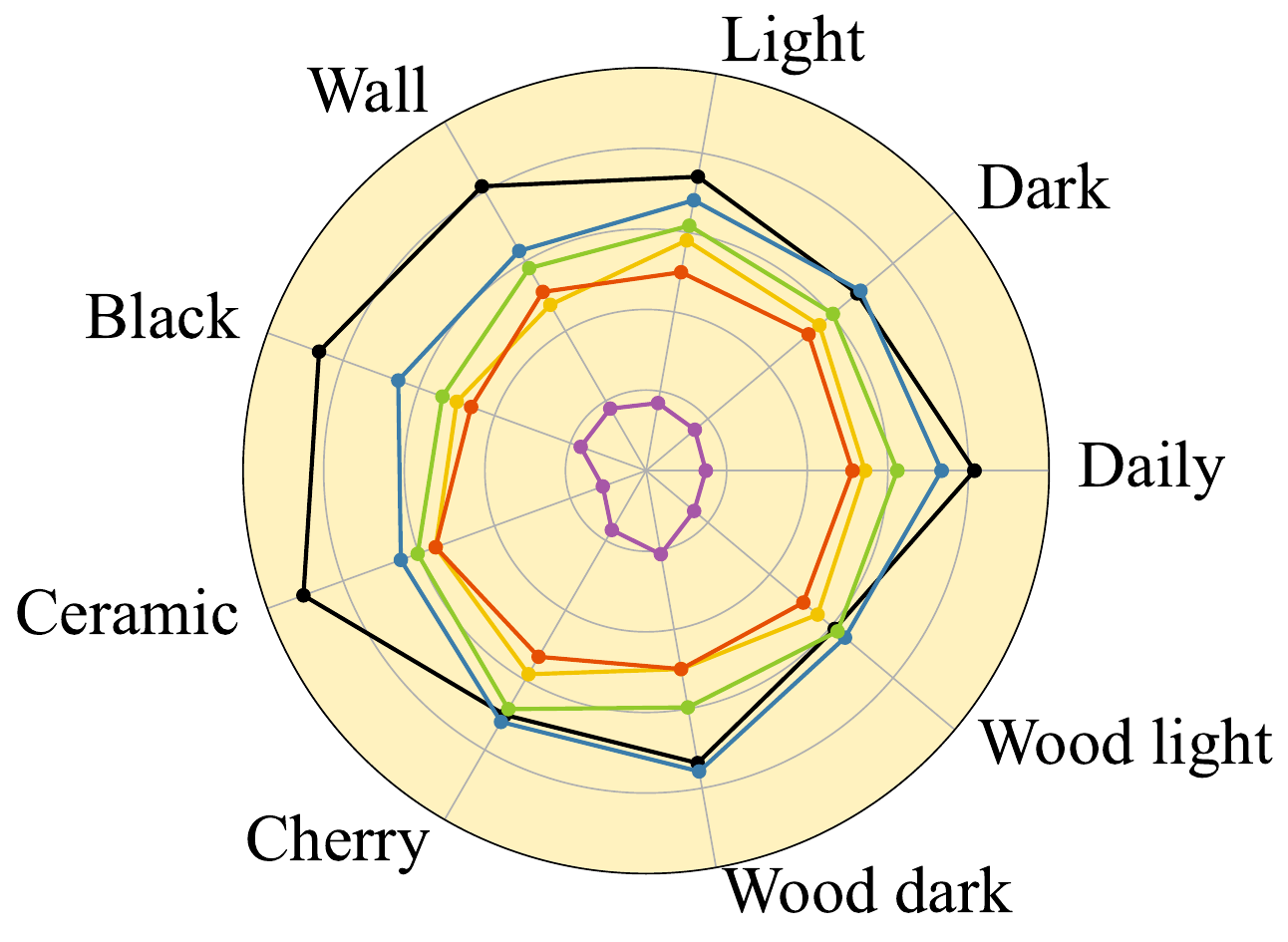}{5mm}%
\subfigoverlap{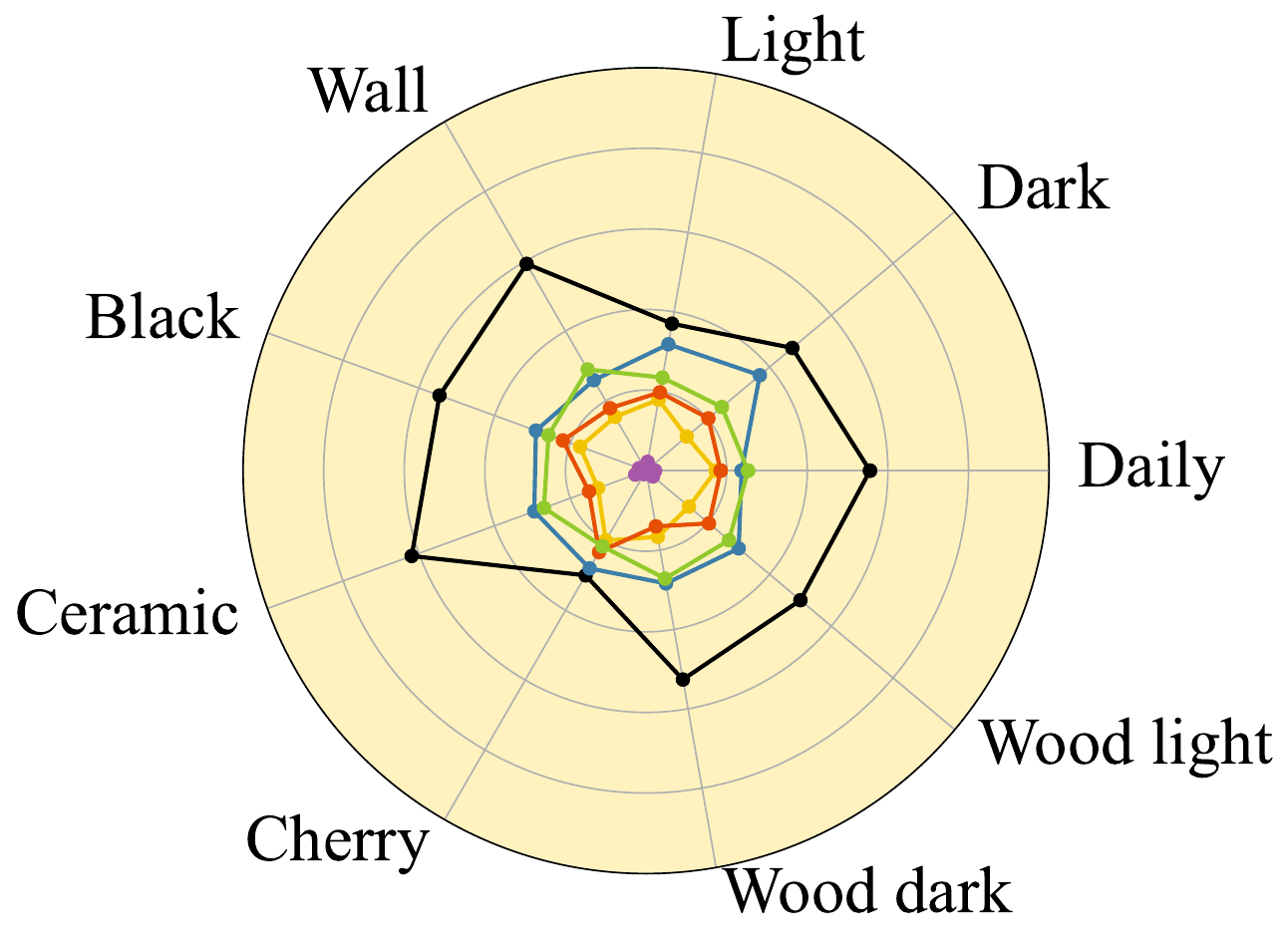}{0mm}

\footnotesize \centerline{\quad\quad\quad\quad   Pi0.5 \hfill OpenVLA \hfill Pi0-Fast \hfill\quad\quad Pi0.5 \hfill OpenVLA \hfill Pi0-Fast   \quad\quad\quad\quad}

\begin{minipage}[]{\linewidth}
  \centering
  \centerline{\includegraphics[width = 0.65\textwidth]{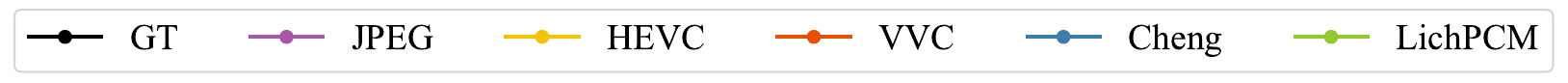}}
\end{minipage}
\vspace{-2mm}

\vspace{-2.5mm}
\caption{Radar map for SR indicator in EmbodiedComp. \CLA{Main object} (Left) and \CLB{Table/Background} (Right) are illustrated separately. For all VLAs, compression leads to SR degradation, while Learned codecs have better fidelity than pixel-level. (zoom-in for detail)}
\vspace{-2mm}
\label{fig:radar}
\end{figure*}

\begin{figure*}
\centering
\begin{minipage}[]{\linewidth}
  \centering
  \centerline{\includegraphics[width = \textwidth]{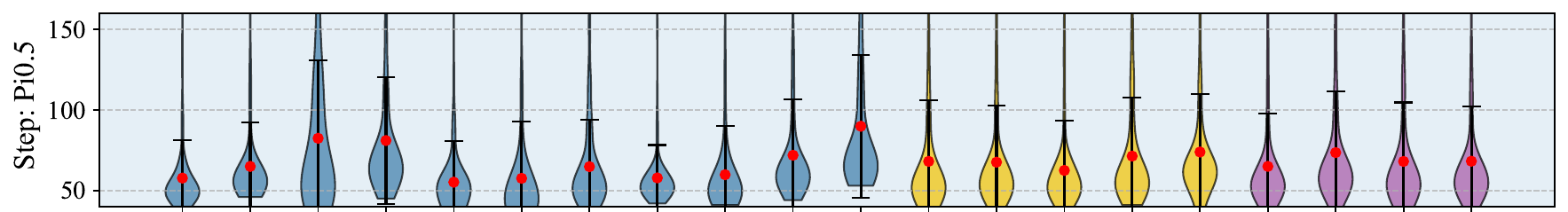}}
\end{minipage}
\begin{minipage}[]{\linewidth}
  \centering
  \centerline{\includegraphics[width = \textwidth]{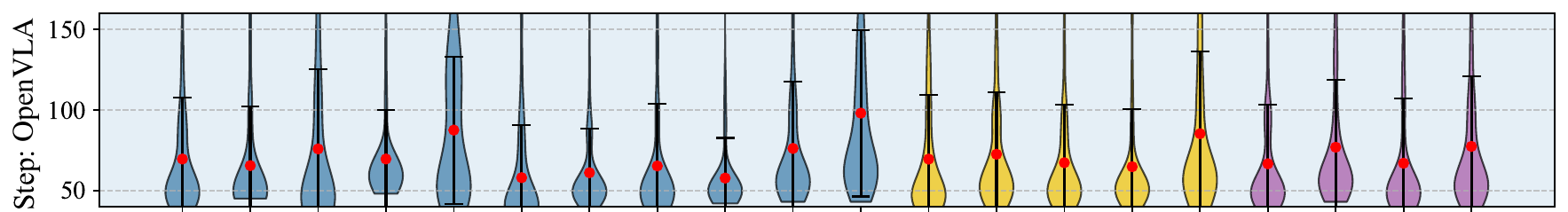}}
\end{minipage}
\begin{minipage}[]{\linewidth}
  \centering
  \centerline{\includegraphics[width = \textwidth]{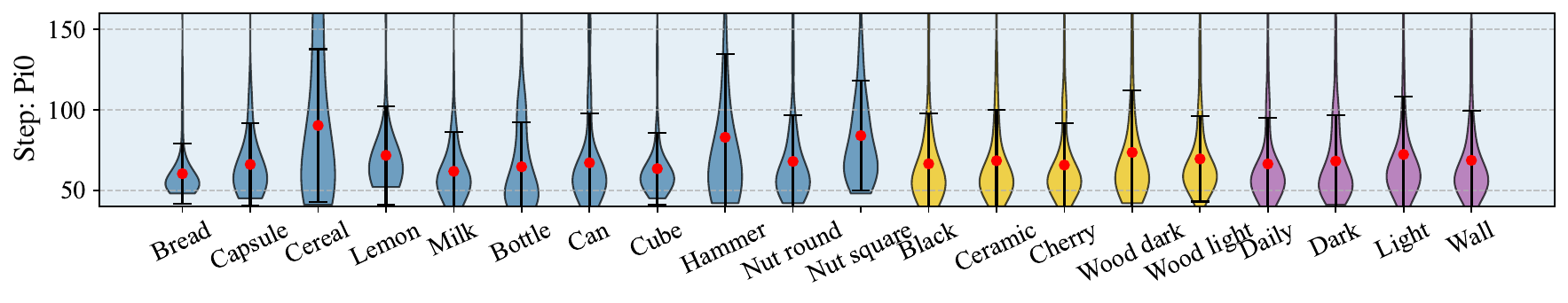}}
\end{minipage}

\vspace{-2mm}

\vspace{-2mm}
\caption{Distribution for Step indicator in EmbodiedComp. Most successful cases require at least 40 steps, while exceeding 150 steps indicates failure. Each VLA has its own unskilled \GroupB{Main object}, while the differences between \GroupD{Table} and \GroupE{Background} are minor.}
\vspace{-3mm}
\label{fig:violin}
\end{figure*}

\subsection{Evaluation Criteria}

Unlike the MVS, whose evaluation relies on Top-1/5/10 accuracy, mAP, mIoU, and other bounding-box or mask-based metrics, the RVS is concerned solely with whether the task is accomplished and how long it takes.  Consequently, EmbodiedComp adopts only two indicators: ($\romannumeral1$) Success Rate (SR): the fraction of scenes in which the command is ultimately satisfied. ($\romannumeral2$) Step: the number of VLA iterations to reach success or exhaust the budget.
These two metrics are theoretically grounded for a closed-loop pipeline.  Each iteration injects compression distortion, while the VLA supplies a finite robustness that can partially cancel it. As illustrated at the bottom of Figure \ref{fig:dataset}, two degradation modes:
\begin{itemize}
    \item Negative-feedback regime (robustness dominates): a single distorted frame misleads the policy, but the main object remains visible; the VLA eventually recovers, succeeding after several extra Steps.
    \item Positive-feedback regime (distortion dominates): even a small pose error propagates across iterations, progressively drifting outside the VLA correction range and causing an irreversible failure.
\end{itemize}
Capturing both regimes requires not only SR but also the Step; hence both metrics are mandatory for an accurate characterization of Embodied AI compression performance.

\section{Data Analysis}

\subsection{The Difference among HVS/MVS/RVS}

Section \ref{sec:ref} argued theoretical perceptual differences among HVS, MVS and RVS; here we supply quantitative evidence from EmbodiedComp. Figure \ref{fig:corr} correlates six HVS-oriented IQA metrics, six MVS-oriented segmentation metrics, and the SR/Step scores of Pi0.5, OpenVLA and Pi0-Fast with bpp across every original and compressed frame (methods indexed in Section \ref{sec:candidates}). Averaged Spearman Rank-order Correlation Coefficient (SRCC) and Pearson Linear Correlation Coefficient (PLCC) is 0.53 for HVS, 0.41 for MVS and below 0.20 for RVS, showing humans are most sensitive to compression, generic vision models less so, and single embodied trials largely uncorrelated with bpp—although average SR eventually tracks bitrate. HVS-oriented compression is thus near saturation, MVS compression is maturing, and RVS-oriented compression remains wide open. Moreover, across different VLAs the correlation stays low, whereas within any single VLA the SR–Step correlation exceeds 0.9, revealing divergent `value systems' among policies; aligning these preferences will be a central challenge for future Embodied codecs.

% Relative Value Table
% The performance degradation percentage of (GT→Normal) and (Normal→Ultra-low) bitrate in EmbodiedComp, and their relative ratio. A higher value indicates that Normal bitrate compression is the dominant factor in quality degradation, while a lower value indicates Ultra-low bitrate.
\begin{table*}[t]
\centering
    \caption{The performance degradation percentage of (GT→Normal) and (Normal→Ultra-low) bitrate in EmbodiedComp, and their relative ratio. A higher ratio indicatesNormal bitrate compression is the dominant factor in quality degradation, which occurs in MVS; while a lower value indicates Ultra-low bitrate dominance, which is for RVS. [Keys: \CLA{Highest}; \CLC{Second Highest}; \CLB{Lowest}; \CLD{Second Lowest}.]}
    \label{tab:main}
    \vspace{-8pt}
    \renewcommand\arraystretch{1.4}
    \belowrulesep=0pt\aboverulesep=0pt
    \resizebox{\linewidth}{!}{
% --- 修正了 \multirow 计数 ---
\begin{tabular}{c:l|c:c:c:c|c:c:c:c:c|c}
\toprule
\multicolumn{2}{c|}{Environment} & \multicolumn{4}{c|}{Background Material Changed} & \multicolumn{5}{c|}{Table Material Changed} & \multirow{2}{*}{Overall} \\
\cmidrule(lr){1-2} \cmidrule(lr){3-6} \cmidrule(lr){7-11}
\multicolumn{1}{l|}{Type} & \multicolumn{1}{c|}{Indicator} & \multicolumn{1}{c|}{Dark} & \multicolumn{1}{c|}{Daily} & \multicolumn{1}{c|}{Light} & \multicolumn{1}{c|}{Walls} & \multicolumn{1}{c|}{Cherry} & \multicolumn{1}{c|}{Black} & \multicolumn{1}{c|}{Wood Dark} & \multicolumn{1}{c|}{Wood Light} & \multicolumn{1}{c|}{Ceramic} &  \\ \midrule
% --- 修改开始 ---
% 1. 所有 \textsubscript{L / E} 的内容已根据新数据更新
% 2. 修正了 HVS-FR 的 \multirow 计数
\multirow{5}{*}{\begin{tabular}[c]{@{}c@{}}HVS\\ -FR\end{tabular}} & PSNR & 9.700\textsubscript{(74.4/7.67)} & 8.811\textsubscript{(73.4/8.33)} & 5.991\textsubscript{(69.6/11.6)} & 3.075\textsubscript{(61.4/20.0)} & 9.506\textsubscript{(73.0/7.68)} & 9.061\textsubscript{(72.1/7.96)} & 3.183\textsubscript{(62.3/19.6)} & 9.132\textsubscript{(73.6/8.06)} & 5.892\textsubscript{(70.7/12.0)} & 3.135\textsubscript{(61.4/19.6)} \\
& SSIM\cite{iqa:ssim} & 3.258\textsubscript{(24.2/7.42)} & 3.055\textsubscript{(19.9/6.52)} & 3.519\textsubscript{(23.0/6.53)} & 3.008\textsubscript{(24.6/8.19)} & 3.341\textsubscript{(18.4/5.50)} & 2.881\textsubscript{(19.3/6.72)} & 2.986\textsubscript{(27.8/9.32)} & 3.632\textsubscript{(19.9/5.49)} & 3.199\textsubscript{(28.9/9.04)} & 3.167\textsubscript{(22.6/7.13)} \\
& LPIPS\cite{iqa:LPIPS} & 2.407\textsubscript{(30.4/12.6)} & 2.440\textsubscript{(27.2/11.2)} & 2.351\textsubscript{(29.3/12.4)} & 2.253\textsubscript{(30.7/13.6)} & 2.821\textsubscript{(28.4/10.1)} & 2.216\textsubscript{(26.7/12.1)} & 2.311\textsubscript{(32.3/14.0)} & 2.894\textsubscript{(28.8/9.94)} & 2.159\textsubscript{(32.1/14.9)} & 2.358\textsubscript{(29.1/12.3)} \\
& DISTS\cite{iqa:DISTS} & 9.225\textsubscript{(52.3/5.67)} & \CLC{15.73}\textsubscript{(52.2/3.32)} & 6.343\textsubscript{(49.0/7.72)} & 5.153\textsubscript{(48.8/9.47)} & \CLC{24.29}\textsubscript{(55.4/2.28)} & 5.098\textsubscript{(48.4/9.50)} & 7.366\textsubscript{(51.7/7.01)} & 11.46\textsubscript{(51.5/4.49)} & 5.565\textsubscript{(49.1/8.82)} & 5.842\textsubscript{(48.4/8.29)}  \\
& PIEAPP\cite{iqa:Pieapp} & 1.748\textsubscript{(31.1/17.8)} & 2.081\textsubscript{(29.0/14.0)} & \CLB{1.865}\textsubscript{(29.8/16.0)} & \CLD{1.986}\textsubscript{(33.4/16.8)} & 1.857\textsubscript{(30.8/16.6)} & 2.111\textsubscript{(28.0/13.3)} & 1.742\textsubscript{(30.2/17.5)} & \CLB{2.147}\textsubscript{(29.0/13.5)} & 1.759\textsubscript{(31.5/17.9)} & 2.109\textsubscript{(28.6/13.6)} \\
\hdashline
% 3. 修正了 HVS-NR 的 \multirow 计数
\multirow{5}{*}{\begin{tabular}[c]{@{}c@{}}HVS\\ -NR\end{tabular}} & CLIPIQA\cite{iqa:CLIPIQA} & 9.649\textsubscript{(35.4/3.67)} & 6.203\textsubscript{(26.6/4.29)} & 7.271\textsubscript{(28.2/3.88)} & \CLC{31.26}\textsubscript{(29.2/0.93)} & 7.045\textsubscript{(26.8/3.80)} & 8.201\textsubscript{(26.7/3.26)} & \CLC{25.91}\textsubscript{(35.4/1.37)} & 8.024\textsubscript{(29.8/3.71)} & 9.864\textsubscript{(29.2/2.96)} & 8.930\textsubscript{(27.4/3.07)} \\
& DBCNN\cite{iqa:DBCNN} & 3.972\textsubscript{(56.3/14.2)} & 2.814\textsubscript{(53.7/19.1)} & 2.888\textsubscript{(54.8/19.0)} & 2.975\textsubscript{(55.0/18.5)} & 3.126\textsubscript{(53.9/17.2)} & 3.379\textsubscript{(56.9/16.8)} & 2.759\textsubscript{(54.7/19.8)} & 2.981\textsubscript{(55.3/18.5)} & 2.887\textsubscript{(54.1/18.7)} & 3.060\textsubscript{(53.7/17.6)} \\
& HyperIQA\cite{iqa:HyperIQA} & 2.539\textsubscript{(47.0/18.5)} & 2.446\textsubscript{(47.5/19.4)} & 2.302\textsubscript{(47.8/20.7)} & 2.430\textsubscript{(47.7/19.6)} & 2.628\textsubscript{(47.9/18.2)} & 2.476\textsubscript{(48.4/19.6)} & 2.274\textsubscript{(47.0/20.7)} & 2.320\textsubscript{(46.7/20.1)} & 2.314\textsubscript{(47.0/20.3)} & 2.393\textsubscript{(46.9/19.6)} \\
& MANIQA\cite{iqa:Maniqa} & 2.089\textsubscript{(43.1/20.6)} & 2.679\textsubscript{(46.6/17.4)} & 2.788\textsubscript{(47.0/16.9)} & 3.593\textsubscript{(46.3/12.9)} & 2.593\textsubscript{(45.0/17.4)} & 3.282\textsubscript{(46.4/14.1)} & 3.066\textsubscript{(46.8/15.3)} & 2.188\textsubscript{(44.0/20.1)} & 2.621\textsubscript{(47.3/18.0)} & 2.665\textsubscript{(44.4/16.7)} \\
& QualiCLIP\cite{iqa:QualiCLIP} & 2.295\textsubscript{(55.6/18.8)} & 2.857\textsubscript{(50.9/17.8)} & 3.161\textsubscript{(53.2/16.8)} & 3.634\textsubscript{(56.5/15.5)} & 2.700\textsubscript{(50.0/18.5)} & 3.386\textsubscript{(55.3/16.3)} & 3.564\textsubscript{(56.0/15.7)} & 3.185\textsubscript{(53.0/16.6)} & 3.620\textsubscript{(55.5/15.3)} & 3.350\textsubscript{(53.5/16.0)} \\
\hdashline
% 4. 修正了 MVS 的 \multirow 计数 (模板有5行数据，新数据有6行，以模板为准)
\multirow{5}{*}{\begin{tabular}[c]{@{}c@{}}MVS\end{tabular}} & SegFormer\cite{seg:SegFormer} & 9.644\textsubscript{(78.7/8.16)} & 7.862\textsubscript{(76.3/9.70)} & 9.806\textsubscript{(77.4/7.90)} & 9.559\textsubscript{(77.8/8.14)} & 6.947\textsubscript{(74.6/10.7)} & 10.82\textsubscript{(78.7/7.27)} & 9.254\textsubscript{(77.1/8.33)} & 9.001\textsubscript{(77.4/8.60)} & 10.27\textsubscript{(79.2/7.72)} & 9.114\textsubscript{(77.5/8.50} \\
& Deeplabv3+\cite{seg:deeplabv3plus} & \CLA{45.50}\textsubscript{(86.7/1.91)} & \CLA{40.34}\textsubscript{(86.5/2.14)} & \CLA{35.50}\textsubscript{(86.1/2.43)} & \CLA{51.17}\textsubscript{(85.8/1.68)} & \CLA{34.48}\textsubscript{(86.3/2.50)} & \CLA{36.49}\textsubscript{(86.7/2.38)} & \CLA{48.44}\textsubscript{(85.9/1.77)} & \CLA{43.87}\textsubscript{(86.2/1.97)} & \CLA{58.71}\textsubscript{(85.9/1.46)} & \CLA{41.66}\textsubscript{(86.3/2.07)} \\
& SegNext\cite{seg:SegNext} & \CLC{15.69}\textsubscript{(76.3/4.86)} & 15.60\textsubscript{(77.6/4.98)} & \CLC{13.03}\textsubscript{(74.4/5.71)} & 13.52\textsubscript{(76.4/5.65)} & 9.901\textsubscript{(75.4/7.61)} & \CLC{16.91}\textsubscript{(74.7/4.42)} & 14.53\textsubscript{(76.4/5.26)} & \CLC{18.62}\textsubscript{(76.4/4.11)} & \CLC{15.56}\textsubscript{(78.1/5.02)} & \CLC{14.32}\textsubscript{(76.3/5.32)} \\
& Swin\cite{seg:Swin} & 11.06\textsubscript{(78.0/7.05)} & 11.61\textsubscript{(79.0/6.80)} & 9.473\textsubscript{(75.8/8.00)} & 13.47\textsubscript{(77.9/5.78)} & 17.50\textsubscript{(80.4/4.59)} & 9.465\textsubscript{(75.9/8.02)} & 10.99\textsubscript{(77.6/7.06)} & 14.39\textsubscript{(78.3/5.44)} & 7.775\textsubscript{(75.9/9.77)} & 11.30\textsubscript{(77.8/6.88} \\
& SETR\cite{seg:SETR} & 7.566\textsubscript{(75.2/9.94)} & 8.234\textsubscript{(76.6/9.31)} & 10.65\textsubscript{(78.8/7.41)} & 8.587\textsubscript{(76.5/8.91)} & 5.073\textsubscript{(70.4/13.9)} & 15.07\textsubscript{(82.1/5.45)} & 8.370\textsubscript{(75.8/9.05)} & 8.344\textsubscript{(76.1/9.12)} & 10.54\textsubscript{(78.8/7.48)} & 8.764\textsubscript{(77.0/8.78)} \\
\hdashline
% 5. 修正了 RVS 的 \multirow 计数
\multirow{6}{*}{RVS} & Pi0-Fast (Step)\cite{vla:pi0} & \CLD{1.588}\textsubscript{(19.1/12.0)} & \CLB{1.322}\textsubscript{(17.6/13.3)} & 2.659\textsubscript{(24.1/9.07)} & \CLB{1.950}\textsubscript{(20.9/10.7)} & 1.811\textsubscript{(20.8/11.5)} & 2.159\textsubscript{(24.3/11.2)} & \CLB{1.608}\textsubscript{(20.8/13.0)} & 2.651\textsubscript{(20.6/7.77)} & \CLB{1.323}\textsubscript{(16.2/12.2)} & \CLB{1.836}\textsubscript{(20.6/11.2)} \\
& Pi0-Fast (SR)\cite{vla:pi0} & 1.609\textsubscript{(25.7/16.0)} & \CLD{1.472}\textsubscript{(24.4/16.6)} & 2.892\textsubscript{(32.6/11.3)} & 2.007\textsubscript{(28.1/14.0)} & 1.867\textsubscript{(28.4/15.2)} & 2.304\textsubscript{(33.2/14.4)} & 1.859\textsubscript{(28.6/15.4)} & 2.502\textsubscript{(27.2/10.9)} & \CLD{1.419}\textsubscript{(21.8/15.3)} & \CLD{1.951}\textsubscript{(27.9/14.3)} \\
& OpenVLA (Step)\cite{vla:openvla} & \CLB{1.460}\textsubscript{(22.6/15.5)} & 3.036\textsubscript{(21.4/7.05)} & 4.120\textsubscript{(24.9/6.03)} & 5.150\textsubscript{(27.6/5.35)} & \CLB{1.561}\textsubscript{(21.8/14.0)} & 12.11\textsubscript{(25.9/2.13)} & 2.014\textsubscript{(23.7/11.8)} & 3.802\textsubscript{(29.7/7.80)} & 4.098\textsubscript{(20.5/5.00)} & 3.053\textsubscript{(24.2/7.93)} \\
& OpenVLA (SR)\cite{vla:openvla}  & 1.719\textsubscript{(28.1/16.3)} & 4.242\textsubscript{(27.7/6.54)} & 5.170\textsubscript{(32.0/6.19)} & 5.749\textsubscript{(33.9/5.90)} & 2.283\textsubscript{(29.6/13.0)} & 10.41\textsubscript{(31.6/3.03)} & 2.172\textsubscript{(31.5/14.5)} & 4.843\textsubscript{(35.9/7.41)} & 6.877\textsubscript{(25.1/3.65)} & 3.764\textsubscript{(30.6/8.13)} \\
& Pi0.5 (Step)\cite{vla:pi05} & 1.838\textsubscript{(35.0/19.0)} & 2.021\textsubscript{(36.3/18.0)} & \CLD{2.030}\textsubscript{(37.7/18.6)} & 2.165\textsubscript{(39.1/18.1)} & \CLD{1.622}\textsubscript{(34.4/21.2)} & \CLB{1.994}\textsubscript{(40.9/20.5)} & \CLD{1.613}\textsubscript{(37.3/23.2)} & \CLD{2.163}\textsubscript{(38.2/17.7)} & 3.631\textsubscript{(34.6/9.54)} & 2.024\textsubscript{(37.2/18.4)} \\
& Pi0.5 (SR)\cite{vla:pi05} & 1.800\textsubscript{(48.2/26.8)} & 2.084\textsubscript{(51.0/24.5)} & 2.156\textsubscript{(53.6/24.9)} & 2.276\textsubscript{(55.2/24.2)} & 1.874\textsubscript{(50.5/27.0)} & \CLD{2.040}\textsubscript{(57.2/28.0)} & 1.667\textsubscript{(52.5/31.5)} & 2.197\textsubscript{(52.9/24.1)} & 3.315\textsubscript{(48.1/14.5)} & 2.050\textsubscript{(52.3/25.0)} \\
% --- 修改结束 ---
\bottomrule
\end{tabular}
    }
\end{table*}

\subsection{The Own Feature of RVS}

This section analyzes in detail the internal relationships of the RVS internal metrics, namely the SR and Step of the three VLAs.
Figure \ref{fig:radar} visualizes SR before and after compression across main object, table, and background instances. At uncompressed Ground Truth (GT), Pi0.5 exceeds 0.9 on every object and attains 1.0 on most; OpenVLA drops uniformly to about 0.8; Pi0-Fast trades accuracy for speed, scoring only 0.4 on the boxed `Milk', and beyond 0.6 on the stable `Bottle'. SR scales with object familiarity—Common, Tools, and Food (Rarely seen), validating the taxonomy in Section \ref{sec:ref}. Cherry-colored tables and high-luminance backgrounds hurt OpenVLA and Pi0-Fast alike. After compression Pi0.5 and Pi0-Fast degrade markedly, whereas the stronger generalization of OpenVLA yields smaller loss and post-compression SR even surpassing Pi0.5. Among codecs, JPEG is totally unacceptable, HEVC/VVC also incur higher degradations, while the two generative methods preserve semantics rather than pixels and retain higher SR. Thus, the factorial coverage of EmbodiedComp across objects, tables and backgrounds exposes the interplay between VLA priors and codec design.

For already succeeded instances, Figure \ref{fig:violin} histograms the Step count. Across all VLAs, success requires $\geq$ 40 iterations; beyond 150 the success probability vanishes. We validated all instances in EmbodiedComp and the empirical maximum is 239 steps, after which every trajectory collapses into an unrecoverable error. We therefore cap the maximum iteration budget at 250. Mean Steps are almost identical among Pi0.5, OpenVLA and Pi0-Fast, with a similar distribution\footnote{Noted the assertion here is in successful samples, Step and SR are irrelevant. But for the entire EmbodiedComp, to ensure fair evaluation, failed samples are penalized up to the maximum steps 250. Thus, Step and SR are still correlated in the experimental results.}, indicating that their performance gaps are driven by success/failure rather than efficiency. Each VLA exhibits a single unskilled object (e.g. Pi0.5: `Lemon'; OpenVLA: `Milk') and all models consume extra steps on `Cereal' and `Nut square'; Table/background choice, however, has negligible effect. Thus, object identity—not scene furnishing—governs step expenditure, underscoring the rationality of diverse object suite in EmbodiedComp.

\section{Experiment}

\subsection{Simulation Settings}

% 在RoboSuite仿真中，我们采取了以下设置，以保证Embodied IQA的100个高质量测试序列能够提供公平的评估。机器人本体选用 UR5e 六自由度机械臂，末端安装Robotiq85夹爪，控制模式为笛卡尔空间位姿控制（OSC-Pose）末端位置与姿态均由 PD 控制。仿真模拟时间步长=0.002 s，控制频率设为 10 Hz。主体物的初始姿态通过 UniformRandomSampler 进行随机采样，范围为桌面中心 ±0.2 m，并与桌面和背景一同渲染，物体排布原则遵照全局SEED生成的唯一seed_chain，使得环境的随机性可以复现。两个相机通过 MuJoCo 的 off-screen renderer来获取图像，分别在Gripper上方10cm，以及固定在机械臂底部，像素为256。有关成功（即跳出循环）的判定条件，EmbodiedComp会在仿真过程中监控当前主体物的空间坐标，我们将Pick/Push定义为7cm的垂直/水平位移，而Press则为简单的物理接触。具体的摩擦、阻尼、相机等参数提供在补充材料中。

To ensure reproducible and fair evaluation, the 100 EmbodiedComp test sequences are generated in Robosuite with the following configuration. A 6-DoF UR5e arm equipped with a Robotiq-85 gripper is commanded in Cartesian space via the OSC-Pose controller; both position and orientation are regulated by PD servos running at 10 Hz with a simulation timestep of 0.002s. The main object is dropped at an (x, y) pose sampled uniformly within $±$0.2m of the table center; table, background and main object layouts are combined under a global SEED to produce a recordable seed chain, ensuring the reproducibility of environmental randomness. Two $256\times256$ RGB cameras—one fixed 10cm above the gripper and one at the robot base—capture off-screen images through MuJoCo renderer. Task success is monitored online: Pick/Push is declared when the centre-of-mass of main object moves 7cm vertically/horizontally; Press is recorded upon detectable contact. Full friction, damping and camera intrinsics are listed in the supplement.

\begin{figure*}
\centering
\begin{minipage}[]{0.24\linewidth}
  \centering
  \centerline{\includegraphics[width = \textwidth]{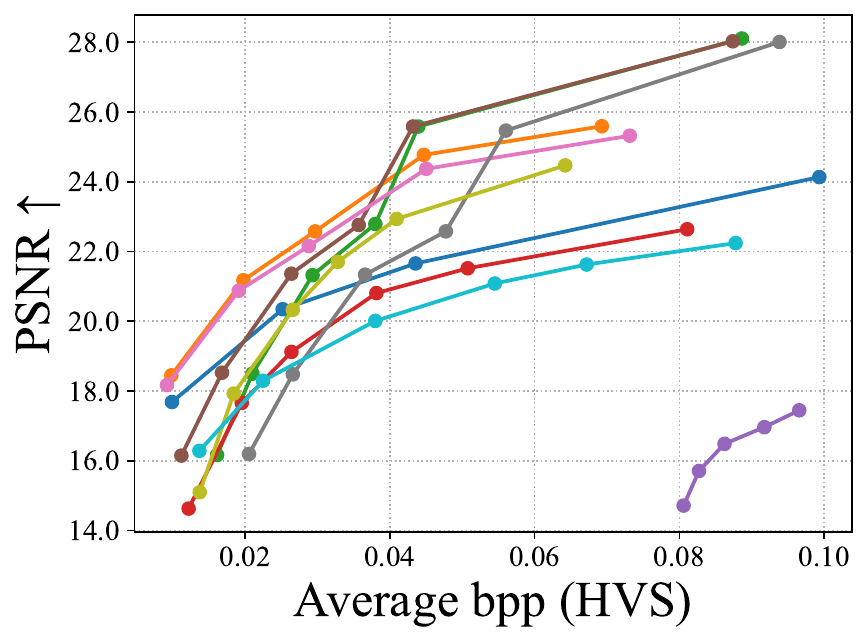}}
\end{minipage}
\vspace{-1mm}
\begin{minipage}[]{0.24\linewidth}
  \centering
  \centerline{\includegraphics[width = \textwidth]{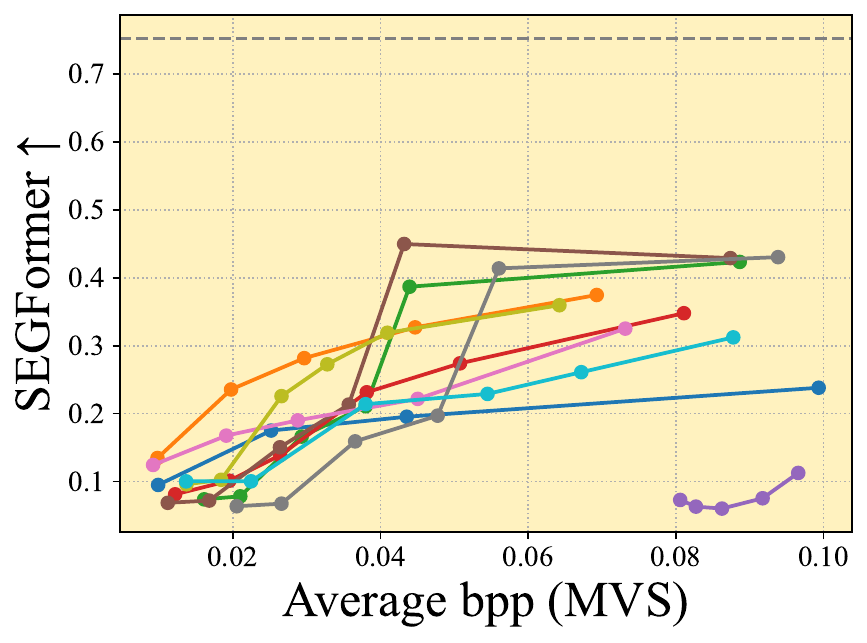}}
\end{minipage}
\vspace{-1mm}
\begin{minipage}[]{0.24\linewidth}
  \centering
  \centerline{\includegraphics[width = \textwidth]{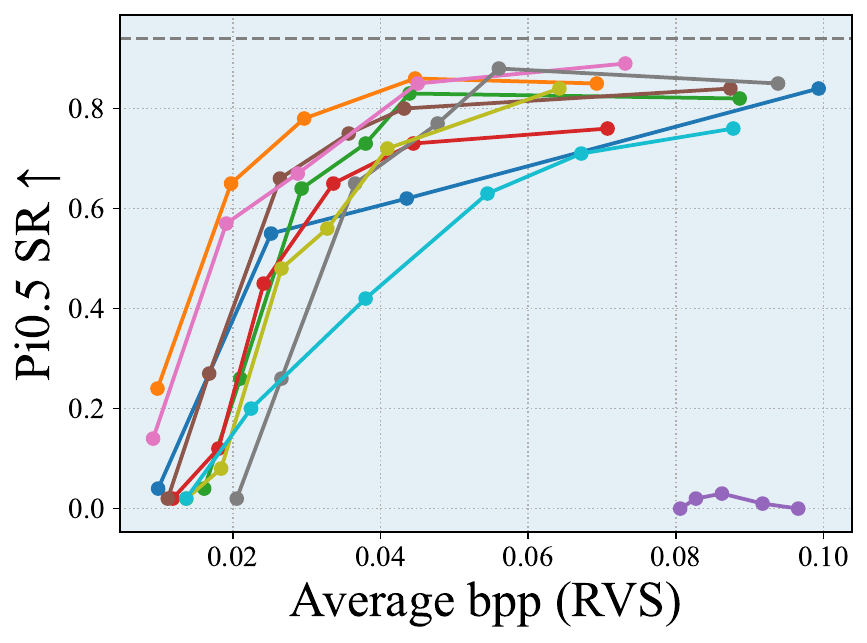}}
\end{minipage}
\vspace{-1mm}
\begin{minipage}[]{0.24\linewidth}
  \centering
  \centerline{\includegraphics[width = \textwidth]{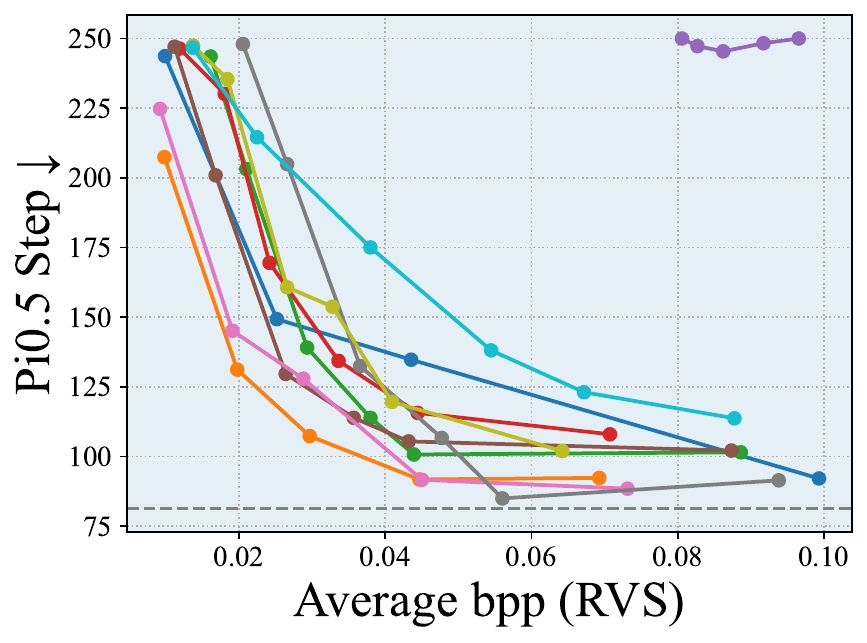}}
\end{minipage}
\vspace{-1mm}

\begin{minipage}[]{0.24\linewidth}
  \centering
  \centerline{\includegraphics[width = \textwidth]{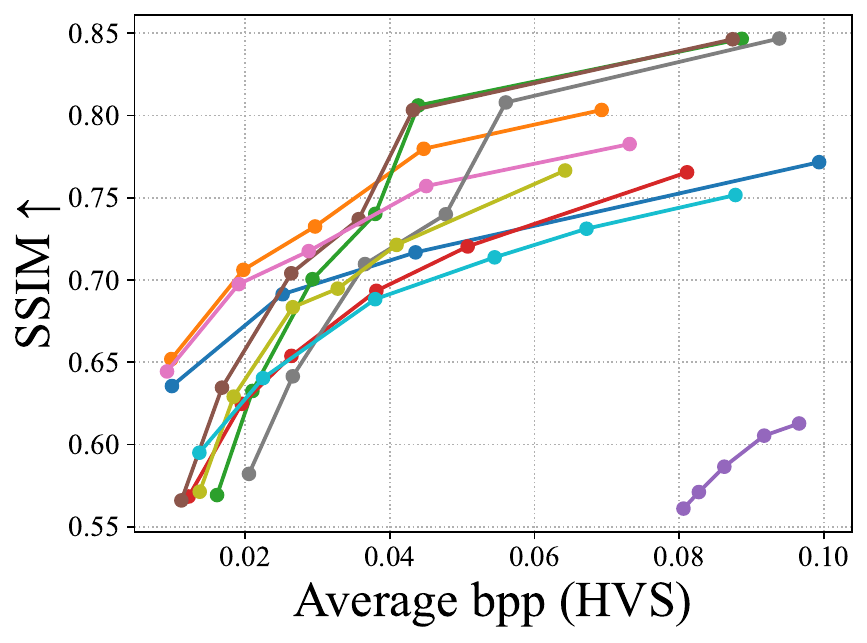}}
\end{minipage}
\vspace{-1mm}
\begin{minipage}[]{0.24\linewidth}
  \centering
  \centerline{\includegraphics[width = \textwidth]{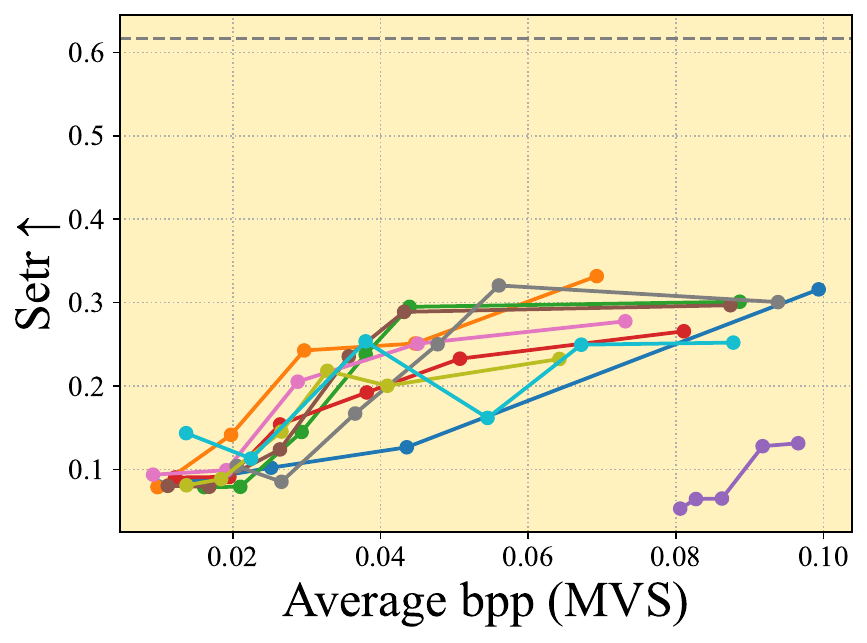}}
\end{minipage}
\vspace{-1mm}
\begin{minipage}[]{0.24\linewidth}
  \centering
  \centerline{\includegraphics[width = \textwidth]{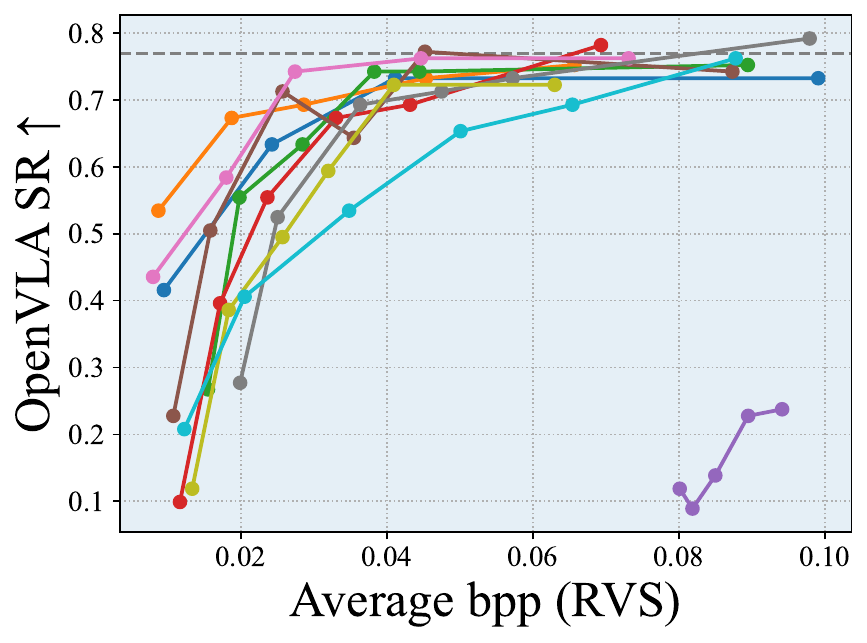}}
\end{minipage}
\vspace{-1mm}
\begin{minipage}[]{0.24\linewidth}
  \centering
  \centerline{\includegraphics[width = \textwidth]{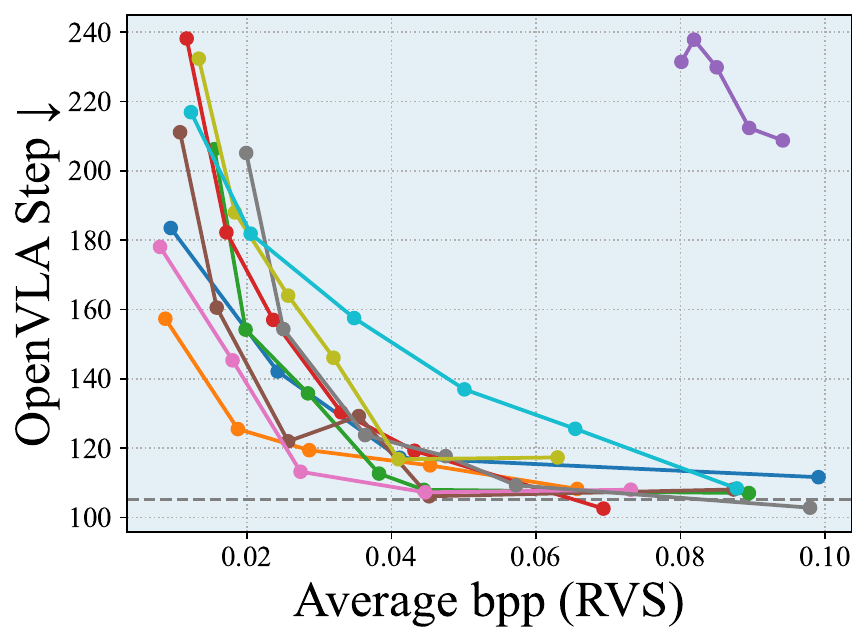}}
\end{minipage}
\vspace{-1mm}

\begin{minipage}[]{0.24\linewidth}
  \centering
  \centerline{\includegraphics[width = \textwidth]{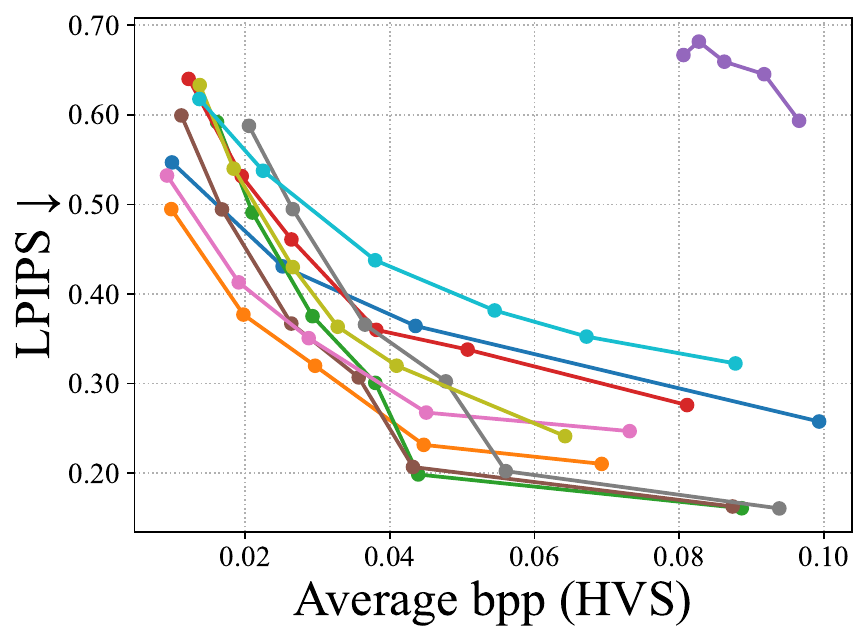}}
\end{minipage}
\begin{minipage}[]{0.24\linewidth}
  \centering
  \centerline{\includegraphics[width = \textwidth]{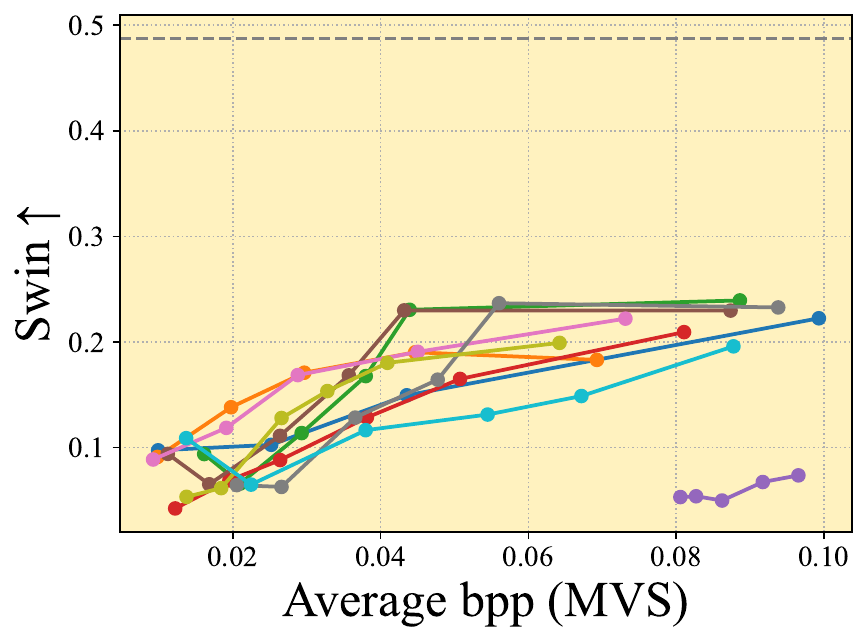}}
\end{minipage}
\begin{minipage}[]{0.24\linewidth}
  \centering
  \centerline{\includegraphics[width = \textwidth]{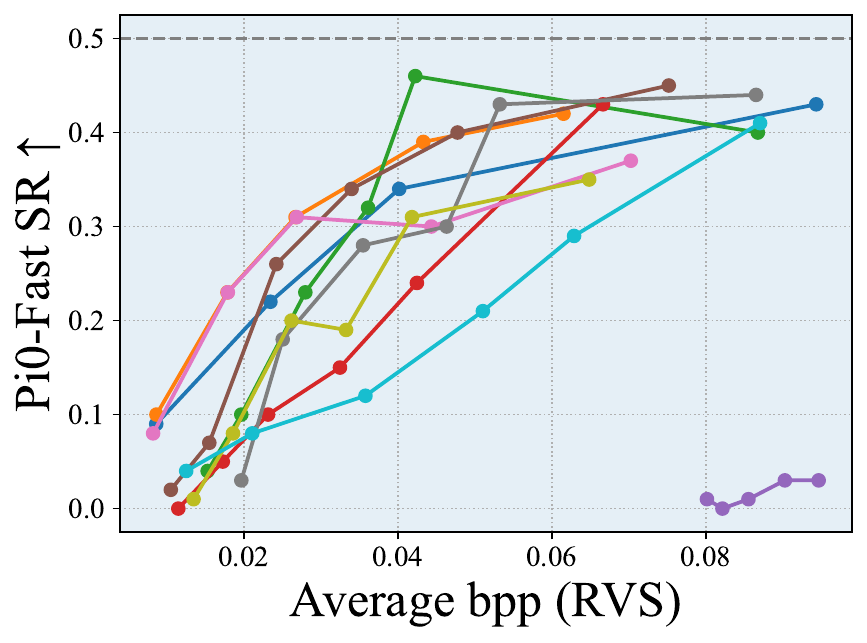}}
\end{minipage}
\begin{minipage}[]{0.24\linewidth}
  \centering
  \centerline{\includegraphics[width = \textwidth]{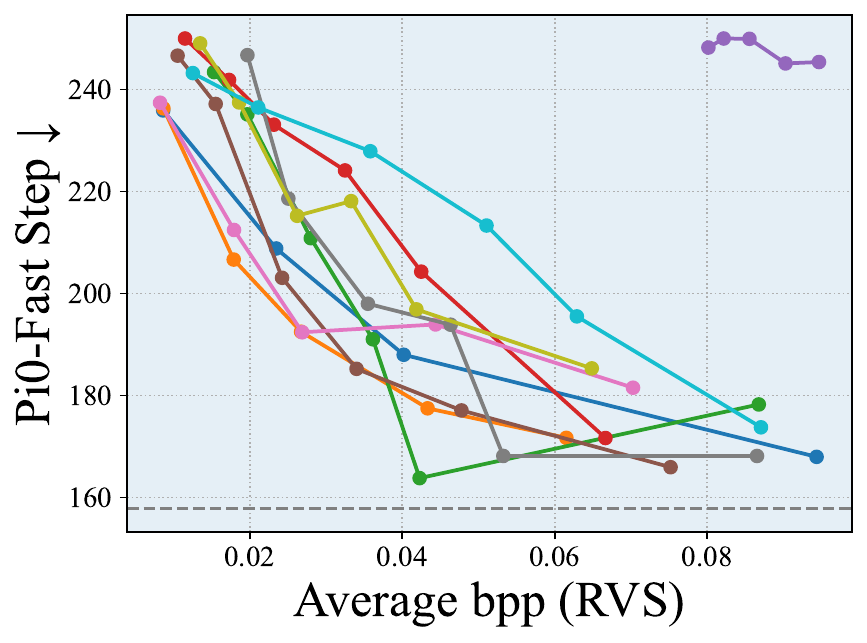}}
\end{minipage}

\begin{minipage}[]{\linewidth}
  \centering
  \centerline{\includegraphics[width = \textwidth]{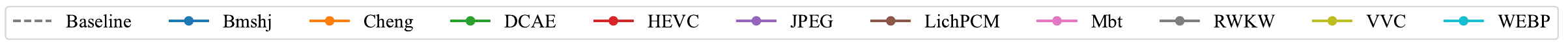}}
\end{minipage}
\vspace{-1mm}

\vspace{-3mm}
% \caption{Rate-Performance curve of \textbf{HVS}/\CLB{MVS}/\CLA{RVS}-oriented indicator in EmbodiedComp. HVS indicator degrades steadily along with bpp; MVS indicator is already low when entering the upper bitrate range of Embodied AI; and RVS firstly remains robust, then significantly drop at 0.05bpp, then becoming unacceptable at 0.015bpp.}
\caption{Rate-Performance curve of \textbf{HVS}/\CLB{MVS}/\CLA{RVS}-oriented indicator in EmbodiedComp. HVS indicator degrades steadily along with bpp; MVS indicator is already low since 0.1bpp; and RVS firstly remains robust, then significantly drop, and finally becomes unacceptable.}
% 结果显示现有的codecs都难以满足Embodied AI在Ultra-low bitrate下的需求，尤其
\vspace{-2mm}
\label{fig:curve}
\end{figure*}

For VLA training, OpenVLA emits a single pose; supervision is the L1 distance between its output and the expert pose. Pi0-Fast and Pi0.5 emit action sequences, so we treat them as trajectories and minimize the L2 flow-matching loss. All models are fine-tuned with LoRA (batch size 32) on 4×NVIDIA H200 NVL 141GB GPUs for 25,000 epochs—sufficient for convergence (Section \ref{sec:train}). For inference, both simulation and Real-world experiment runs on an NVIDIA GeForce RTX 5090 32GB GPU for graphics.

\subsection{Benchmark Candidates}
\label{sec:candidates}

We comprehensively implement three codec types\footnote{Classic Learned codec refers to CompressAI, and the latest Learned codecs are proposed after 2025 to ensure superiority on HVS/MVS.}, including ($\romannumeral1$) Pixel-level: HEVC \cite{codec:hevc}, JPEG \cite{codec:jpeg}, VVC \cite{codec:vvc}, WEBP \cite{codec:webp}; ($\romannumeral2$) Classic end-to-end: Bmshj \cite{codec:bmshj}, Cheng \cite{codec:cheng}, Mbt \cite{codec:mbt}; and ($\romannumeral3$) The latest end-to-end: DCAE \cite{codec:dcae}, LichPCM \cite{codec:lichpcm}, RWKV \cite{codec:rwkv}. Following the modeling of IoT and channels in Section \ref{sec:dis}, we define the working range of Embodied AI as 0.015 to 0.1 bpp, with the lower/upper limits representing Ultra-low/Normal bitrates respectively.

All of these codecs are evaluated using indicators, including ($\romannumeral1$) HVS Full-Reference (FR): PSNR, SSIM\cite{iqa:ssim}, LPIPS\cite{iqa:LPIPS}, DISTS\cite{iqa:DISTS}, PieAPP\cite{iqa:Pieapp}; ($\romannumeral2$) HVS No-Reference (NR): CLIPIQA\cite{iqa:CLIPIQA}, DBCNN\cite{iqa:DBCNN}, HyperIQA\cite{iqa:HyperIQA}, ManIQA\cite{iqa:Maniqa}, QualCLIP\cite{iqa:QualiCLIP}; ($\romannumeral3$) MVS: SegFormer\cite{seg:SegFormer}, Deeplabv3+\cite{seg:deeplabv3plus}, SegNext\cite{seg:SegNext}, Swin\cite{seg:Swin}, SETR\cite{seg:SETR}; and ($\romannumeral4$) RVS: our trained VLA models.

\subsection{Experiment Result and Discussions}

% 在宏观上，我们先把所有压缩统一视为Distorted image，并研究其与Reference的性能差异。Table 2 把 EmbodiedComp 基准里“GT→Normal”与“Normal→Ultra-low”两段码率降级所造成的性能跌幅（\%）并排列出，并算出二者的相对比值（括号内“前者/后者”）。比值越高，说明Normal 码率的轻微压缩才是质量主因；比值越低，则表明Ultra-low 码率的极端压缩才是罪魁祸首。可以明显看出，MVS在全部十个比值上都是最大的，普遍 >70，最高达 86.7（DeepLabV3+），意味着只要把图像从 GT 压到 Normal，性能就掉七成以上，再压到 Ultra-low 只是雪上加霜，Normal 压缩是主导因素。RVS 类，则在八个比值上最小，两个萨和嗯次小。比值骤降到 20–40 区间，例如 OpenVLA (SR) 仅 28.0，Pi0.5 (Step) 35.0；跌幅主力发生在 Ultra-low 段，说明机器人视觉对轻微压缩几乎无感，直到码率被砍到极端才明显失灵。HVS 指标夹在中间，比值 25–55 不等。
% 环境维度上，同一类型指标在不同材质下走势一致，只是绝对跌幅略有浮动，表明码率效应远大于纹理差异，也验证了表格设计的普适性。
% 总结，尽管Embodied AI在广义上也是一种机器，但它和general机器的区别，甚至比机器和人还大。这种在Ultra-low bitrate上的性能暴跌，是未来Embodied image compression需要解决的关键因素。

At the macro level we treat every compressed frame as Distorted images and measure the drop relative to the Reference. Table \ref{tab:main} juxtaposes the degradation (\%) induced by the bpp range in EmbodiedComp: `GT→Normal' versus `Normal→Ultra-low', and reports their ratio (former/latter). High ratio means that mild Normal-rate loss already dominates the final score; while a low ratio flags Ultra-low rate as the true culprit.
MVS ratios are the highest on all ten codecs, consistently exceed 30 and peaking at 41.66 for DeepLabV3+; performance collapses as soon as GT is pushed to Normal, and Ultra-low only adds marginal extra damage. RVS ratios are the smallest in eight cases and second-smallest in the remaining two, plunging to the 1–5 band (e.g., 1.951/1.836 for Pi0-Fast SR/Step); the major drop occurs only when the rate is driven to the Ultra-low cliff, confirming that robotic vision is almost indifferent to light compression. HVS-oriented metrics sit in between, with ratios scattered from 2 to 10. Across table materials the same curve shape holds, merely shifting absolute, proving that bitrate, not texture, governs the trend.
In short, \CLA{although Embodied AI is a special type of `machines', its divergence from general machine vision is even larger than the gap between machine and human.} 
%The abrupt RVS failure at Ultra-low bitrate is the pivotal problem that future Embodied image compression must solve.

% At the micro level, 我们详细对比了每种codec在各种bpp上的性能。
% Figure 7 给出的是 EmbodiedComp 基准上HVS,MVS,RVS(SR)和RVS(Step)指标随平均bpp变化的 Rate–Performance 曲线，横坐标统一为Embodied AI工作的bitrate区间，纵坐标是各自任务对应的性能得分。codecs整体走势清晰揭示出三类视觉系统的“抗压”差异：（1）HVS从 0.10 bpp 开始性能最高，随 bpp 降低呈单调、近似线性下滑；到 0.02 bpp 时跌幅已超 50 \%，说明人眼对压缩的感知较为均匀；（2）MVS在0.1bpp就已经很低，这表明轻微压缩已让分割模型丢失关键纹理边缘，Embodied AI bitrate区间带来的压缩只是“补刀”，与 Table 2 的“高比值”结论互为印证；（3）RVS前段（0.10→0.06 bpp）几乎水平横走，性能下降不到 5\%，显出强鲁棒；进入 0.04 bpp 附近出现第一次明显拐点，得分开始跳水；到 0.02 bpp 时曲线最陡，性能跌至不可接受区间。这种“先稳后崩”的形状量化验证了：VLA 模型对轻度压缩不敏感，但会在某点开始剧烈下降。考虑到现实的Embodied AI应用中，虽然不一定存在信道最恶劣且设备最多的0.015bpp情况，但不可能永远在0.1bpp的理想上限，因此这个断崖下跌需要解决。
%我们在同一张 Rate–Performance 即图7里，把十种编/解码器拉到一起，显示面向HVS/MVS/RVS曲线的高低排序基本一致，但细看起鲁棒性仍能分出梯队。

At the micro level, Figure \Ref{fig:curve} plots per-codec Rate–Performance curves for HVS, MVS, RVS(SR) and RVS(Step) across the bpp range relevant to embodied transmission. HVS scores are highest at 0.10 bpp and fall almost linearly, losing 50\% by 0.02 bpp, indicating uniform human sensitivity. MVS is already low at 0.10 bpp—light compression erases the texture edges required for segmentation—so merely works at the Embodied AI bitrate window, corroborating the high GT→Normal ratios in Table \ref{tab:main}. RVS curves stay flat from 0.10 to 0.06 bpp (5\% drop), then kink sharply around 0.04 bpp and plummet to unusable levels at 0.02 bpp, confirming the `robust-then-cliff' behavior. Although real deployments of Embodied AI yet rarely hit the worst 0.015 bpp, it will seldom enjoy 0.10 bpp at perfect channel condition and abundant bandwidth, \CLA{mitigating this sudden collapse is the critical design target for Embodied image compression.}

\begin{table}[t]
\centering
    % \caption{Absolute robustness comparison between \textit{GPT-4o} and \textit{human} (\textbf{left/right}). Evaluated by 3 tasks, 3 strength, 7 steps, and 7 groups. As the R-bench champion, \textit{GPT-4o} still lags behind \textit{human} across the board.  \CLB{Orange}/\CLA{Blue} denote \textit{GPT-4o} performance below 90\% or above 98\% of \textit{humans}.}
    \caption{Validation for generalization ability for VLAs, to operate unseen objects (o1-o6 listed at the bottom) outsides EmbodiedComp. Their performance align with the original SR and Steps.}
    \label{tab:generalization}
    \vspace{-8pt}
    \renewcommand\arraystretch{1.4}
    \belowrulesep=0pt\aboverulesep=0pt
    \resizebox{\linewidth}{!}{
\begin{tabular}{l|c:cccccc:c}
\toprule
VLA     & Original & \multicolumn{1}{:c}{o1}  & \multicolumn{1}{c}{o2} & \multicolumn{1}{c}{o3} & \multicolumn{1}{c}{o4}   & \multicolumn{1}{c}{o5}  & \multicolumn{1}{c:}{o6} & Mean   \\ \midrule
Pi0.5   & 0.94~    & 0.71~ & 0.94~ & 0.85~   & 1.00~ & 1.00~ & 1.00~  & 0.91~  \\
        & 81~      & 131~  & 63~   & 103~    & 68~   & 85~   & 102~   & 92~    \\ \cdashline{1-9}
OpenVLA & 0.77~    & 0.78~ & 1.00~ & 0.75~   & 0.72~ & 0.75~ & 0.67~  & 0.78~  \\
        & 105~     & 107~  & 53~   & 122~    & 155~  & 144~  & 176~   & 126~   \\ \cdashline{1-9}
Pi0-Fast     & 0.50~    & 0.35~ & 0.29~ & 0.54~   & 0.49~ & 0.60~ & 0.27~  & 0.42~  \\
        & 158~     & 189~  & 222~  & 152~    & 148~  & 120~  & 224~   & 176~   \\ \bottomrule
\end{tabular}}
\vskip\medskipamount   % 表格与图之间的垂直间距
\noindent
\begin{minipage}{0.16\linewidth}
  \centering
  \includegraphics[width=\linewidth]{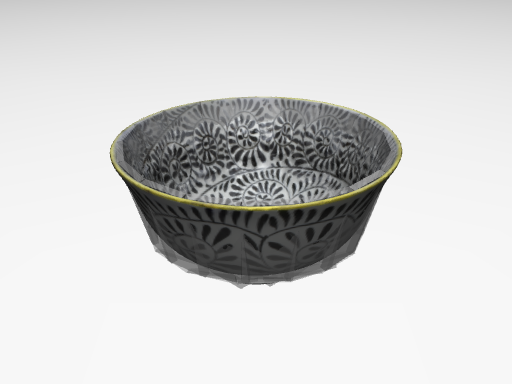}\\[2pt]
%  \footnotesize o1 
\end{minipage}\hfill
\begin{minipage}{0.16\linewidth}
  \centering
  \includegraphics[width=\linewidth]{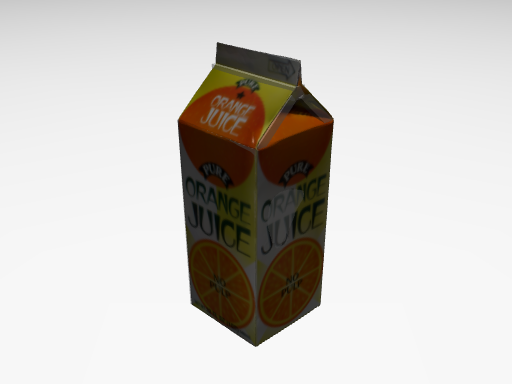}\\[2pt]
%  \footnotesize o2
\end{minipage}\hfill
\begin{minipage}{0.16\linewidth}
  \centering
  \includegraphics[width=\linewidth]{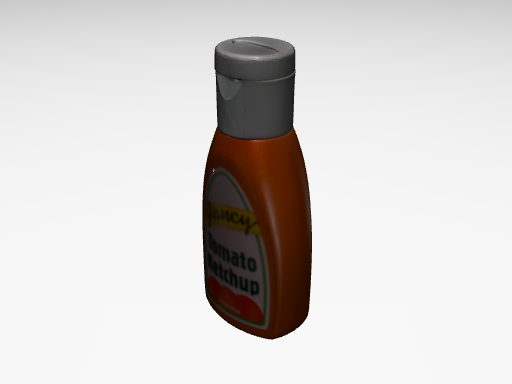}\\[2pt]
%  \footnotesize o3
\end{minipage}\hfill
\begin{minipage}{0.16\linewidth}
  \centering
  \includegraphics[width=\linewidth]{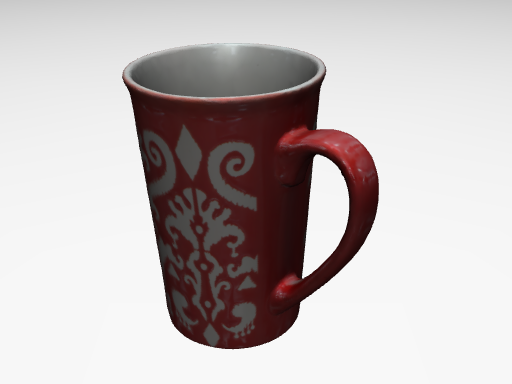}\\[2pt]
%  \footnotesize o4
\end{minipage}\hfill
\begin{minipage}{0.16\linewidth}
  \centering
  \includegraphics[width=\linewidth]{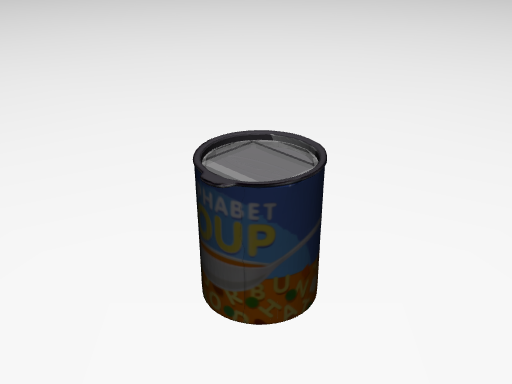}\\[2pt]
%  \footnotesize o5
\end{minipage}\hfill
\begin{minipage}{0.16\linewidth}
  \centering
  \includegraphics[width=\linewidth]{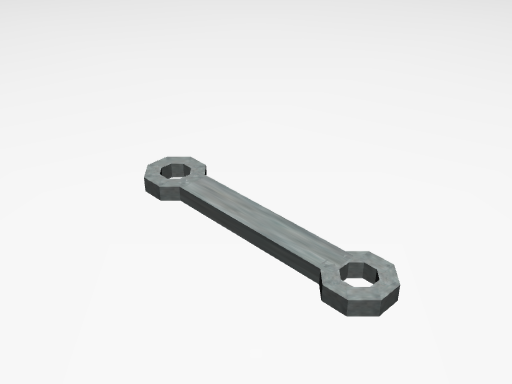}\\[2pt]
%  \footnotesize o6
\end{minipage}
\vspace{-5mm}
\end{table}
Across codec families we observe a clear ranking shift. For HVS, latest-learned codecs dominate, while traditional-learned and pixel-level (e.g., VVC) tie; for MVS, latest-learned retains the lead and traditional-learned surpasses pixel-level codecs because semantic learning helps segmentation, yet for RVS the order inverts—although latest-learned codecs (e.g. DCAE, LichPCM) excel in HVS/MVS, they ultimately underperform traditional-learned codecs (e.g. Cheng). This reversal suggests that the advanced learned models over-fit to HVS/MVS priors and fail to generalize to RVS perception dynamics. In conslusion, \CLA{future codecs targeting Embodied AI must therefore guard against such prior over-fitting to avoid the paradox} that `more advanced' yields `more inferior'.
% 此外，有关pixel-level, traditional-learned，和latest-learned方法，我们观察到如下结果。对于HVS，理所当然是latest-learned最强，而traditional-learned和pixel-level并驾齐驱；对于MVS，latest-learned仍最有优势，此时由于learned codec学习了语义信息，即使是最先进的pixel-level(如VVC)也不如traditional-learned；而RVS则发生了有趣的现象，尽管DCAE与LichPCM这种latest-learned在前面展现了卓越的性能，它们反而打不过traditional-learned codec(如Cheng)。这可能是因为它们过度学习了HVS/MVS的先验，而RVS的感知机理完全不同。宗上，未来的codec若想服务与Embodied AI，必须防止过拟合，避免上述‘越先进，越落后’的现象。

\subsection{Validation for Generalization \& Real-world}

%本文的一切发现还需要建立在以下两个前提（1）VLA执行失败或走额外步骤，是因为压缩的影响，而不是触发了其本身的缺陷；（2）在真实世界中，压缩会对完整的Embodied AI链路造成影响。因此，我们额外增加了Generalization和Real-world两个部分的实验。

All the findings above are based on the following two premises: ($\romannumeral1$) VLA execution failure or additional steps are due to the effect of compression, rather than triggering its inherent defects;  ($\romannumeral2$) In the Real-world, compression will affect the complete Embodied AI pipeline. Therefore, we added two additional validation experiments.

\begin{figure}
\centering
\begin{minipage}[]{0.38\linewidth}
  \centering
  \centerline{\includegraphics[width = \textwidth]{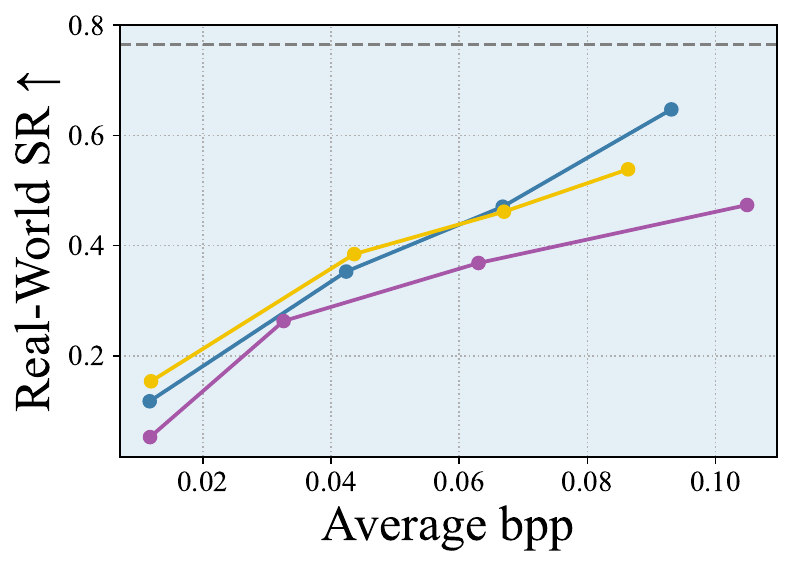}}
\end{minipage}
\begin{minipage}[]{0.38\linewidth}
  \centering
  \centerline{\includegraphics[width = \textwidth]{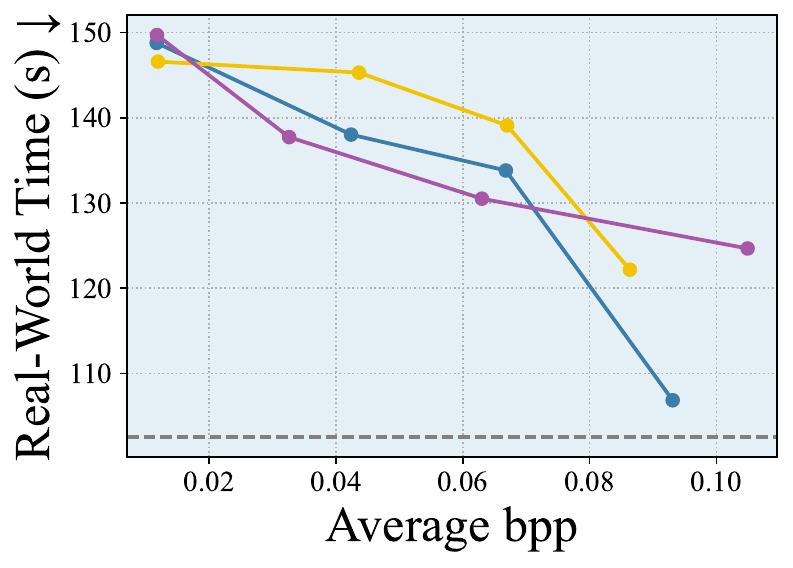}}
\end{minipage}
\begin{minipage}[]{0.18\linewidth}
  \centering
  \centerline{\includegraphics[width = \textwidth]{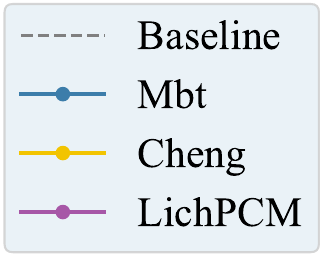}}
\end{minipage}

\begin{minipage}[]{\linewidth}
  \centering
  \centerline{\includegraphics[width = 0.95\textwidth]{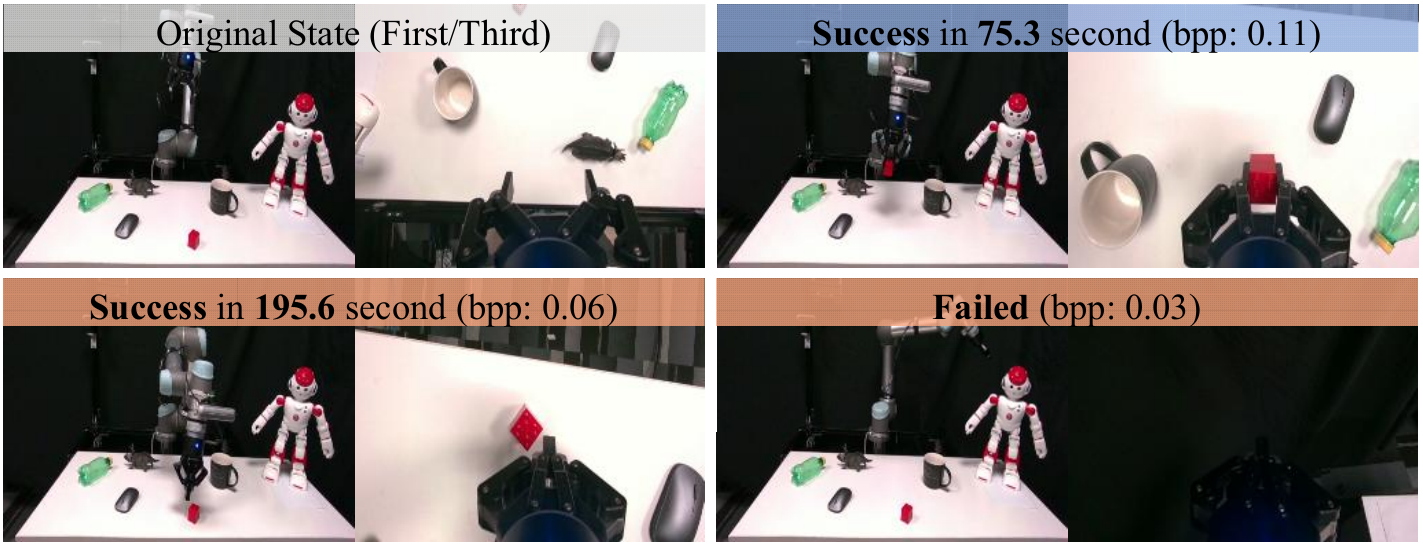}}
\end{minipage}
\vspace{-1mm}

\vspace{-1mm}
\caption{Validation for the Real-world. The Rate-Performance curve align with simulation where both extra time and wrong execution exist, indicating the rationality of EmbodiedComp.}
\vspace{-6mm}
\label{fig:real}
\end{figure}

For generalization, Table \ref{tab:generalization} selects six new items (o1-o6) in RoboSuite, as the main object to be operated. To control variables, the rendering mechanism of table and background remains unchanged. Overall, VLAs never experience a significant performance degradation when processing these new objects (OpenVLA even improved), only incurring a few more steps. This is because these objects are all rigid and share certain characteristics with EmbodiedComp, such as o2 for `Milk' and o3 for `Bottle'. Thus, the performance degradation of VLA during compression is not due to limitations in VLA generalization causing the main object to be identified as an unmanageable new object, but entirely due to distortion caused by compression.
% 总体来看，VLA在处理这些新物体时性能没有明显下降（OpenVLA甚至上升了），只是多花费了一些Step。这是因为以上物体均为刚性，和EmbodiedComp有一定共性，例如o2之于milk，o3之于bottle。因此VLA在压缩时的性能下降，并不是VLA generalization的限制，导致main object被辨认为无法操作的新物体，而完全是来自压缩带来的失真。

For Real-world, we implement the trained Pi0.5 (Blue curve in Figure \ref{fig:train}) into UR5 robotic arm and Robotiq 2F-140 gripper, with a working radius of 85cm. Other settings and evaluation ceireria align with the simualtion, where we used the measurements from a ruler\&stopwatch to represent the SR\&Step in the simulation.
As shown in Figure \ref{fig:real}, we consider the three most representative compression algorithms in the simulation and used them to perform a pick task on 17 types of main objects, executing them a total of 765 times. The trend of the rate-performance curve is consistent with that of the simulation, and the `more advanced' yields `more inferior' event also occurred for LichPCM. Therefore, the sim2real results are reliable, and the relationship between Embodied AI and codec in the real world follows the pattern in  EmbodiedComp.

\section{Conclusion}

Aiming at the evolution from generic machines to Embodied AI in ICM, we propose a novel task:  Embodied Image Compression, to better enable codecs to serve Embodied AI. First, we establish the  EmbodiedComp benchmark, evaluating the codec within a closed-loop based on the characteristics of RVS. Second, we demonstrate the necessity of using codecs in  Embodied AI and define its operating bitrate range. Finally, we validate the performance of advanced codecs, result shows a critical inflection point within the aforementioned range, rendering existing VLAs ineffective. We sincerely hope that EmbodiedComp can inspire better ICM and promote the application of Embodied AI.

\clearpage
% \maketitlesupplementary
\appendix

\section{Limitation and Broader Impact}

Visual-signal compression research is governed by a single non-negotiable precondition: \textbf{the downstream system must already deliver reliable performance when supplied with uncompressed imagery.} Over the past decades the community has moved through four successive targets—(i) pure signal fidelity in the early 2000s, (ii) human subjective preference once high-resolution televisions and tablets became ubiquitous, (iii) machine vision after segmentation and detection networks surpassed 99\% frame-level accuracy, and (iv) most recently MLLMs as they emerged since 2023. Embodied Image Compression is no exception: it can only quantify the `additional' degradation introduced by bitrate reduction once the Embodied agent itself is demonstrably competent.
This axiom gives rise to two explicit limitations of the present work. 

First, our benchmark currently covers manipulation tasks exclusively and does not address navigation. The rationale is empirical: modern VLAs already achieve 80\% success on tabletop primitives in previously unseen office scenes, thereby satisfying the `uncompressed competence' gate. In contrast, the best published VLN policies still attain only 20–30\% success when asked to walk end-to-end to a previously unseen target in a photorealistic indoor environment; \textbf{compressing an already failing policy would merely document an expected collapse rather than reveal codec-specific failure modes.} Equally important, human-activity statistics show that upper-limb manipulation accounts for roughly 60\% of daily interactive behaviour, whereas lower-limb+torso+sensorimotor adjustments contribute the remaining 40\%, so manipulation is presently the higher-impact domain. We will extend EmbodiedComp to navigation once VLN accuracy crosses the same usability threshold that manipulation has already achieved.

Second, the benchmark does not embrace every VLA published to date. We deliberately restrict validation to three representative models—Pi0.5 (highest single-task accuracy), Pi0-Fast (lowest inference latency), and OpenVLA (largest community uptake and open-weight availability).  This decision mirrors the long-standing convention in Image-Compression-for-Machine research, where exactly three downstream tasks (classification, detection, segmentation) are deemed sufficient to characterize a codec.  Smaller-parameter VLAs such as Octo or RT-1 do not yet generalize reliably on our desktop scenes even without compression, \textbf{thereby violating the foundational precondition stated above.} To keep the benchmark aligned with the Embodied AI community, we will periodically refresh the downstream triplet—e.g., adding Pi-0.6, Gemini-Embodied, or future SOTA models—whenever their uncompressed accuracy surpasses that of the current incumbents.

For broader impact, by shifting compression targets from human or generic-machine perception to the Robotic Visual System, this work establishes the first benchmark that couples bitrate reduction with closed-loop manipulation success. The EmbodiedComp dataset and evaluation protocol will guide codec designers toward algorithms that postpone the 0.04bpp `cliff' where VLAs suddenly fail, directly improving bandwidth-limited multi-robot deployments in warehouses, hospitals and homes. Conversely, by exposing how small calibration shifts can reverse codec rankings, we alert the community to avoid over-fitting learned compressors to human or segmentation priors, fostering safer and more reliable cloud-edge collaborative systems.

\section{Simulation Settings}

We conduct manipulation experiments in the MuJoCo 3.3.4 back-end of RoboSuite 1.5.1.  
A 6-DoF UR5e arm equipped with a Robotiq-85 two-finger gripper is commanded in Cartesian space (OSC-pose) at 10 Hz; the simulation step is 0.002s.  
State integration uses explicit Euler; contacts are solved with Newton’s method (tol = 1e-8) and an elliptic friction-cone approximation. Default soft-contact parameters are kept (solref = [0.02, 1], solimp = [0.9, 0.95, 0.001]).  
\begin{itemize}
    \item Solver limits: 100 main iterations, 50 line-search iterations (tol = 0.01), 50 CCD iterations (tol = 1e-6), 10 SDF iterations, 40 SDF initial samples. Gravity is 9.81 $\rm m\cdot s^{-2}$.
    \item Scene: a 0.8 × 0.8 × 0.05 m table whose top is at z = 0.8 m (friction: 1.0, 5e-3, 1e-4). Objects are randomly chosen per episode from according to Section 4. The object is dropped from a uniformly random (±0.2 m, ±0.2 m) position 5 mm above the tabletop.
    \item An episode lasts 300 control steps (30 s): the first 50 steps apply a zero-velocity dummy action to let the object settle; the agent then has 250 steps (25 s) to succeed.
    \item Rendering: off-screen OpenGL, $256\times256$ RGB, $45^{\circ}$ vertical FOV, near = 0.01m, far = 3m, MSAA off. every episode uses a unique seed derived from a global seed.
\end{itemize}
Meanwhile, due to the different output policy of OpenVLA (1 action) and Pi-Series (16 actions), we apply different training loss.
Here, OpenVLA is trained by minimizing the L1 distance between predicted and expert action chunks, with the loss defined as: 
\begin{equation}
    \mathcal{L}_{\text{L1}}=\frac{1}{B\,T\,D}\sum_{b=1}^{B}\sum_{t=1}^{T}\sum_{d=1}^{D}\bigl|A_{b,t,d}-\hat{A}_{b,t,d}\bigr|,
\end{equation}
where \(B\) denotes the mini-batch size, \(T\) denotes the action-chunk horizon (the time steps count in one prediction), \(D\) denotes the dimensionality of a single action vector, \(A_{b,t,d}\) denotes the ground-truth action collected from demonstrations, and \(\hat{A}_{b,t,d}\) denotes the action predicted by the policy head. 
Pi0-Fast and Pi0.5 employ conditional flow matching and optimize the L2 distance between the learned vector field and the analytic target field of a linear Gaussian probability path, with the practical batch loss written as:
\begin{equation}
    \mathcal{L}_{\text{FM}}=\frac{1}{B\,H\,D}\sum_{b=1}^{B}\sum_{h=1}^{H}\sum_{d=1}^{D}\bigl(v_{b,h,d}-u_{b,h,d}\bigr)^{2},
\end{equation}
where  \(H\) denotes the chunk length (equivalent to the horizon \(T\) above), \(v_{b,h,d}=f_{\theta}(o,A^{\tau},\tau)\) denotes the vector field predicted by the action head, and \(u_{b,h,d}=\varepsilon-A\) denotes the ground-truth vector field derived from the path definition \(q(A^{\tau}|A)=\mathcal{N}\!\bigl(\tau A,(1-\tau)\mathbf{I}\bigr)\) with interpolated action \(A^{\tau}=\tau A+(1-\tau)\varepsilon\) and standard Gaussian noise \(\varepsilon\sim\mathcal{N}(\mathbf{0},\mathbf{I})\), while \(\tau\in[0,1]\) denotes the interpolation time sampled uniformly for each training example.

\begin{figure}[t]
    \centering
    \includegraphics[width=0.97\linewidth]{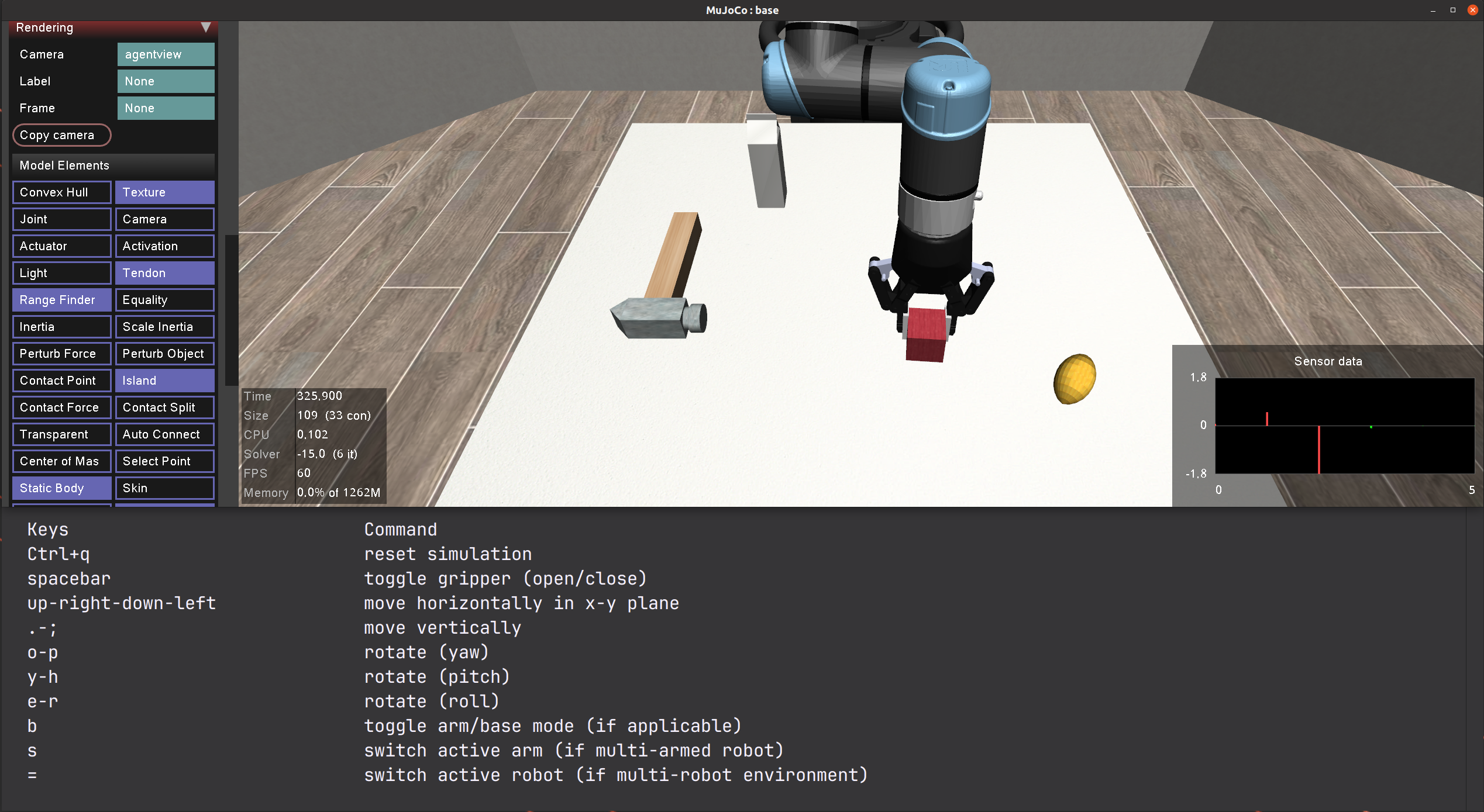}
    \caption{Data collect in simulation environment}
    \label{fig:sim-collect}
\end{figure}

\section{Subjective Data Collection}
The datasets used for fine-tuning VLA are divided into simulation environments and the UR5 real robot. Data collection in the simulation is based on a modified `pick/push/press' task environment in Figure \ref{fig:sim-collect}, with grasping status monitored through a table-top camera view rendered in real time. The position of end-effector of the robotic arm $(x, y, z, r, p, y)$ and the open/closed state of the gripper $g$ are controlled using an Xbox controller or keyboard. Data including joint angles, two camera streams (first and third-person view), and action (position deltas).

\begin{figure}[t]
    \centering
    \includegraphics[width=\linewidth]{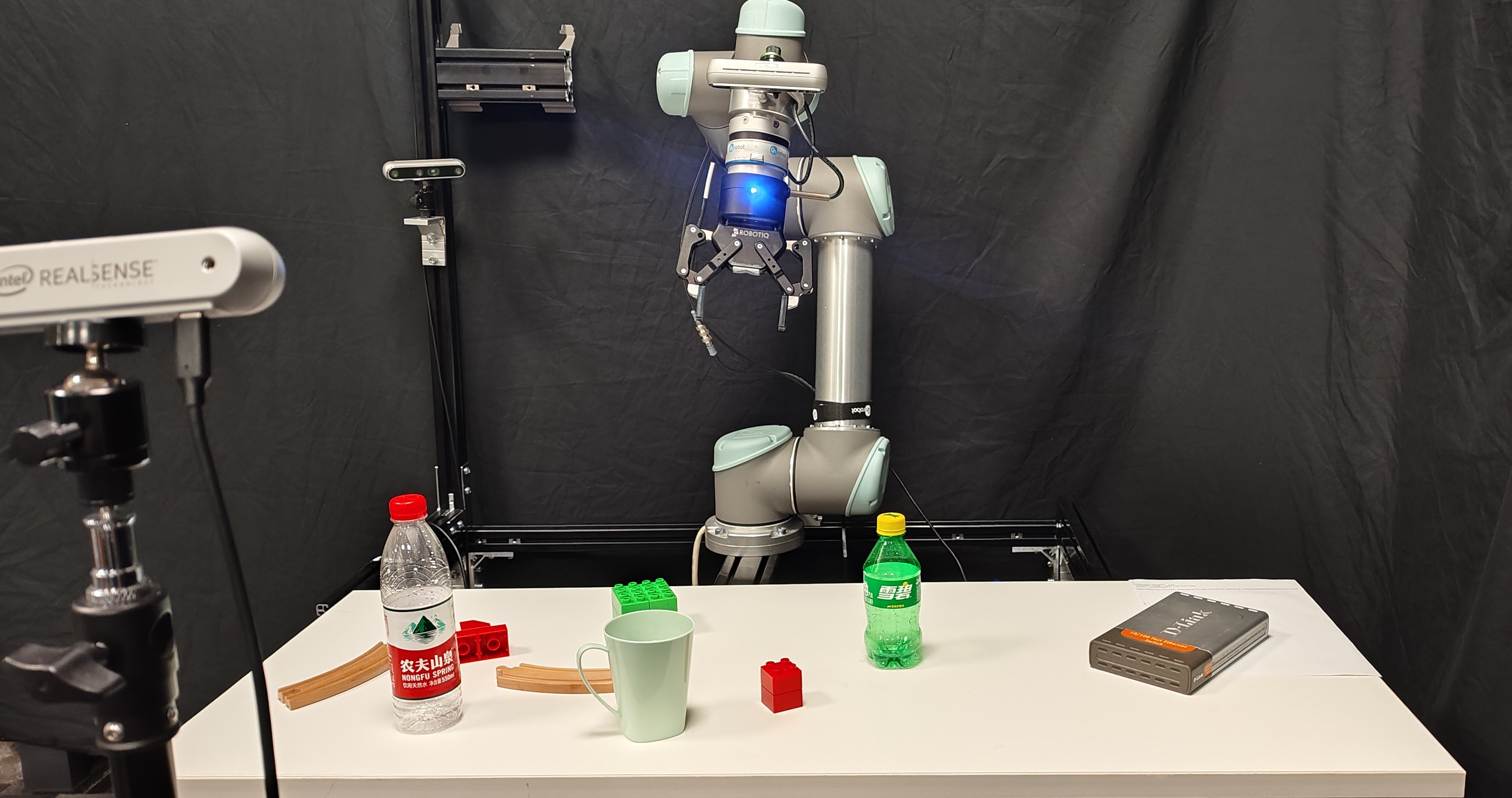}
    % \begin{minipage}{0.3\textwidth}
    %     \includegraphics[width=\textwidth]{supp/supp-real-world2.png}
    % \end{minipage}
    \caption{Data collect in real-world}
    \label{fig:real-collect}
\end{figure}

For Real-world data collection in Figure \ref{fig:real-collect}, an UR5 robotic arm equipped with a Robotiq gripper is used within a custom experimental setup. After setting the UR5 to free-move mode via its control panel, human operator manually moves the arm to perform tasks. Joint angles, two camera streams (wrist view and third-person view, captured by two Intel realsense cameras), and actions (end-effector position deltas) are recorded at 10 Hz frequency and packaged for storage. All datasets are transformed into Lerobot or RLDF format to meet the needs of different VLA. 

The task to perform is textual-based and is pre-defined before the manipulation, completed by the author team.
Then, five experts with experience in Robotic research project participated in the manipulation annotation, labeling 2,000 high-quality simulation trajectories, and 400 Real-world gripper position series, to provide ground truth in VLA training process.
Every expert trajectory is reviewed by at least one additional expert to detect idiosyncratic or sub-optimal motion patterns; samples that deviate from the agreed-upon shortest, collision-free path are discarded and replaced by newly collected demonstrations. Tasks whose wording or object placement may have elicited the anomalous behaviour are re-annotated (instruction text revised or scene reset) and re-recorded until unanimous approval is reached. All data collection adheres to the Declaration of Helsinki and is covered by a signed open-data agreement.

\section{Manipulation Object Description}

This Section introduce all main objects involved in EmbodiedComp. With the format (SR baseline, Step baseline) → (SR compressed, Step compressed), including:
\begin{figure}[!h] 
\centering
\begin{minipage}[t]{0.14\textwidth}
    \vspace{0pt}  
    \includegraphics[width=\textwidth]{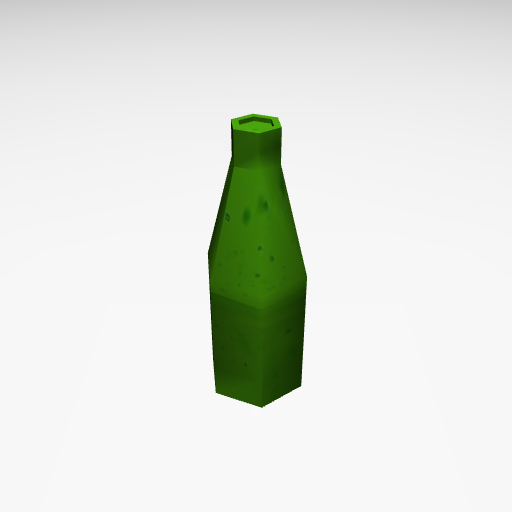}
\end{minipage}
\hfill
\begin{minipage}[t]{0.32\textwidth}
    \vspace{0pt}
    \textbf{Bottle} (0.89,85.7)→(0.61,113.6)\\
    A green glass bottle with a narrow neck suitable for grasping, approximately 18 cm in length, representing a typical bottle-shaped object.
\end{minipage}
\vspace{-1.2mm}
\end{figure}
\begin{figure}[!h] 
\centering
\begin{minipage}[t]{0.14\textwidth}
    \vspace{0pt}  
    \includegraphics[width=\textwidth]{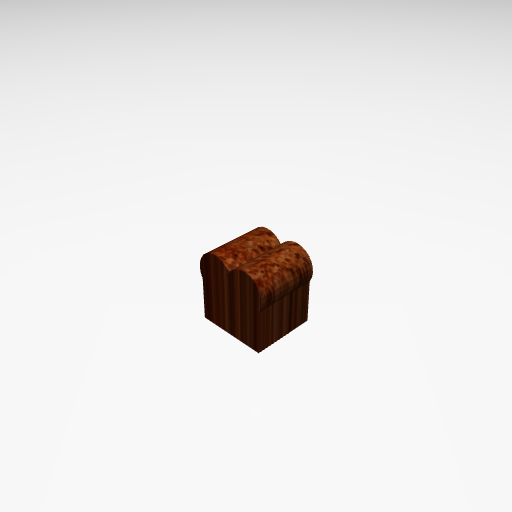}
\end{minipage}
\hfill
\begin{minipage}[t]{0.32\textwidth}
    \vspace{0pt}
    \textbf{Bread} (0.93,69.9)→(0.54,149.7)\\
    A brown, square-shaped bread whose main differences from a cube lie in its raised top and surface texture, approximately 6 cm in length.
\end{minipage}
\vspace{-1.2mm}
\end{figure}
\begin{figure}[!h] 
\centering
\begin{minipage}[t]{0.14\textwidth}
    \vspace{0pt}  
    \includegraphics[width=\textwidth]{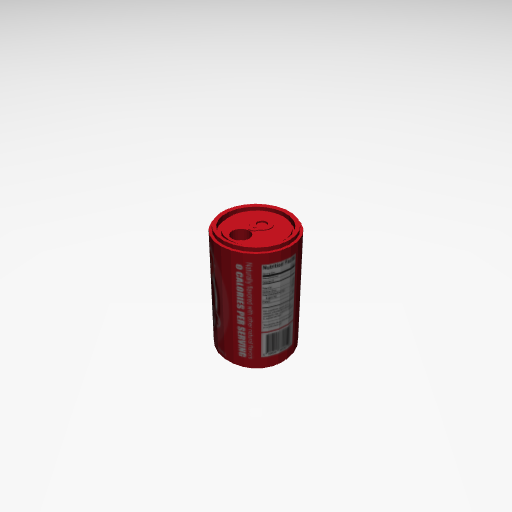}
\end{minipage}
\hfill
\begin{minipage}[t]{0.32\textwidth}
    \vspace{0pt}
    \textbf{Can} (0.83,85.7)→(0.60,137.7)\\
    A classic red cola can with a diameter of approximately 6 cm, representing most canned beverages in daily use.
\end{minipage}
\vspace{-1.2mm}
\end{figure}
\begin{figure}[!h] 
\centering
\begin{minipage}[t]{0.14\textwidth}
    \vspace{0pt}  
    \includegraphics[width=\textwidth]{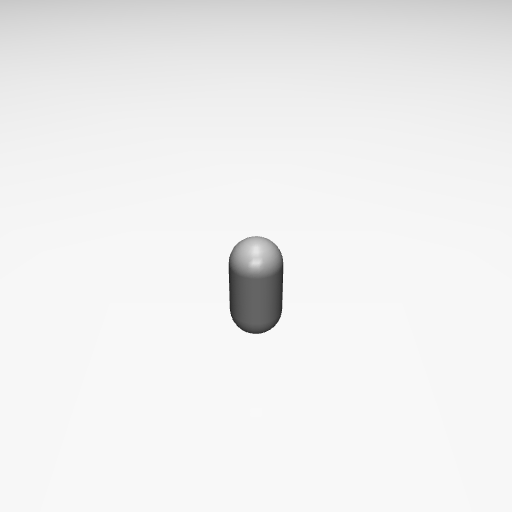}
\end{minipage}
\hfill
\begin{minipage}[t]{0.32\textwidth}
    \vspace{0pt}
    \textbf{Capsule} (0.48,159.4)→(0.28,198.2)\\
    A capsule approximately 3 cm in length with a smooth and small surface, requiring precise positioning for manipulation. It rolls away easily with even a slight touch.
\end{minipage}
\vspace{-1.2mm}
\end{figure}
\begin{figure}[!h] 
\centering
\begin{minipage}[t]{0.14\textwidth}
    \vspace{0pt}  
    \includegraphics[width=\textwidth]{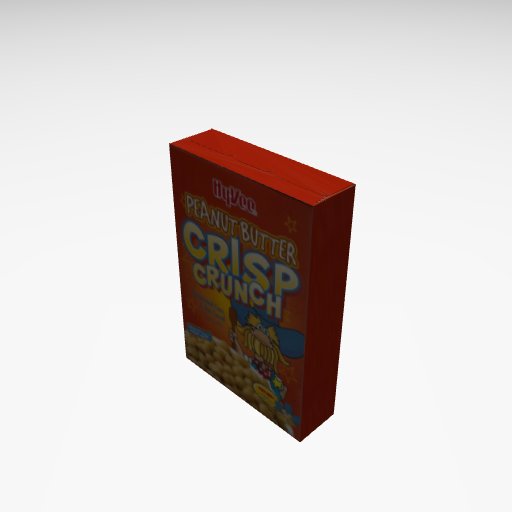}
\end{minipage}
\hfill
\begin{minipage}[t]{0.32\textwidth}
    \vspace{0pt}
    \textbf{Cereal} (0.54,150.1)→(0.29,199.3)\\
    A cereal box approximately 16 cm in length, with a smooth surface and an unstable center of gravity, easily toppled when touched, representing the many rectangular boxed items.
\end{minipage}
\vspace{-1.2mm}
\end{figure}
\begin{figure}[!h] 
\centering
\begin{minipage}[t]{0.14\textwidth}
    \vspace{0pt}  
    \includegraphics[width=\textwidth]{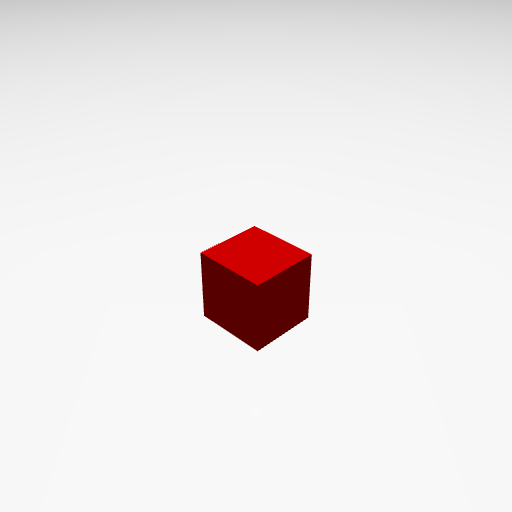}
\end{minipage}
\hfill
\begin{minipage}[t]{0.32\textwidth}
    \vspace{0pt}
    \textbf{Cube} (0.89,75.5)→(0.54,148.9)\\
    A simple red cube with a rough surface and minimal texture, whose structure is easily subjected to stress.
\end{minipage}
\vspace{-1.2mm}
\end{figure}
\begin{figure}[!h] 
\centering
\begin{minipage}[t]{0.14\textwidth}
    \vspace{0pt}  
    \includegraphics[width=\textwidth]{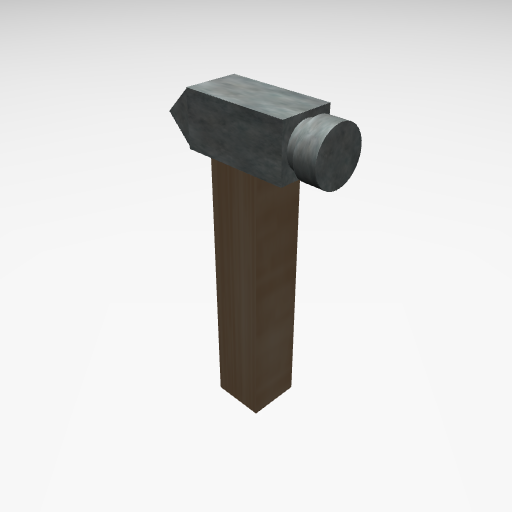}
\end{minipage}
\hfill
\begin{minipage}[t]{0.32\textwidth}
    \vspace{0pt}
    \textbf{Hammer} (0.61,127.2)→(0.39,177.5)\\
    A hammer approximately 20 cm in length with a 4 cm wide handle, featuring an unbalanced center of gravity,representing heavier tools commonly encountered in daily life.
\end{minipage}
\vspace{-1.2mm}
\end{figure}
\begin{figure}[!h] 
\centering
\begin{minipage}[t]{0.14\textwidth}
    \vspace{0pt}  
    \includegraphics[width=\textwidth]{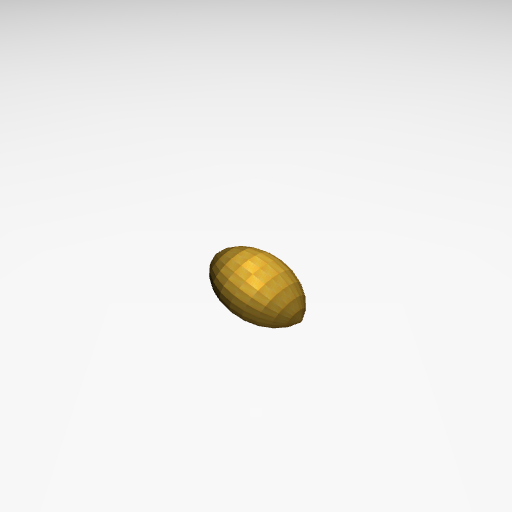}
\end{minipage}
\hfill
\begin{minipage}[t]{0.32\textwidth}
    \vspace{0pt}
    \textbf{Lemon} (0.76,111.5)→(0.40,178.6)\\
    A typical yellow lemon with a smooth surface and an oval shape, requiring grasping from one side and easy to rolling away, representing fruit-like objects.
\end{minipage}
\vspace{-1.2mm}
\end{figure}
\begin{figure}[!h] 
\centering
\begin{minipage}[t]{0.14\textwidth}
    \vspace{0pt}  
    \includegraphics[width=\textwidth]{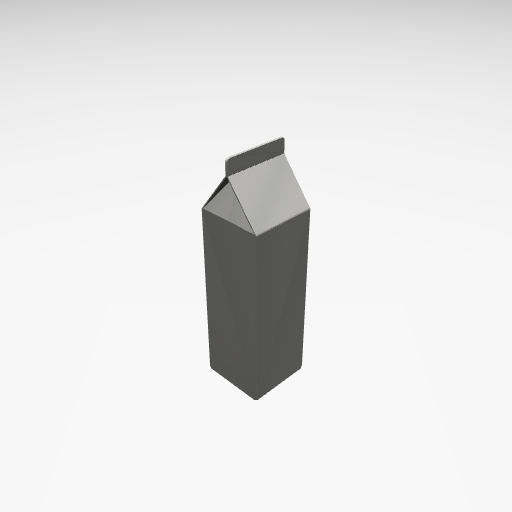}
\end{minipage}
\hfill
\begin{minipage}[t]{0.32\textwidth}
    \vspace{0pt}
    \textbf{Milk} (0.89,85.1)→(0.44,169.9)\\
    A common white paper milk carton, approximately 19 cm tall, with a simple texture but easily toppled, representing typical boxed liquids.
\end{minipage}
\vspace{-1.2mm}
\end{figure}
\begin{figure}[!h] 
\centering
\begin{minipage}[t]{0.14\textwidth}
    \vspace{0pt}  
    \includegraphics[width=\textwidth]{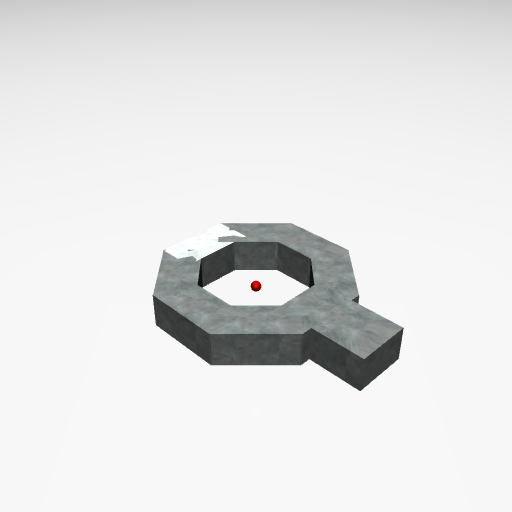}
\end{minipage}
\hfill
\begin{minipage}[t]{0.32\textwidth}
    \vspace{0pt}
    \textbf{RoundNut} (0.76,114.7)→(0.40,178.5)\\
    An aluminum round nut featuring a central round hole and a short protrusion, relatively thin and requiring edge grasping.
\end{minipage}
\vspace{-1.2mm}
\end{figure}
\begin{figure}[!h] 
\centering
\begin{minipage}[t]{0.14\textwidth}
    \vspace{0pt}  
    \includegraphics[width=\textwidth]{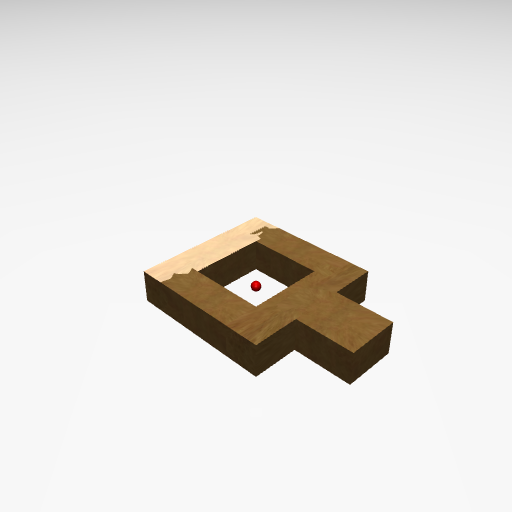}
\end{minipage}
\hfill
\begin{minipage}[t]{0.32\textwidth}
    \vspace{0pt}
    \textbf{SquareNut} (0.63,148.8)→(0.32,198.5)\\
    A wooden square nut featuring a central square hole and a short protrusion, relatively thin and requiring edge grasping.
\end{minipage}
\vspace{-1.2mm}
\end{figure}
% 综上，EmbodiedComp对主体物的考量较为全面；既由简单/困难的基线难度；又有对压缩敏感/鲁棒的样本。因此，它对真实世界中操作物体的外观、材质、受力结构进行了全面的表征，可以客观地量度压缩对VLA带来的影响。

In summary, EmbodiedComp provides a comprehensive consideration of main objects; it offers both easy/hard baseline difficulty and compression-sensitive/robust samples. Therefore, it comprehensively characterizes the appearance, material, and stress structure of manipulated objects in the Real-world, and can objectively measure the negative impact of compression on VLA.

% \begin{figure*}[tb]
%     \centering
%     \includegraphics[width = \textwidth]{vis/Feature-12.pdf}

%     \caption{Low-level feature probability distribution of Mo-QAD, visualized by 12 different MLLMs as source decoder. Different colors denote \GroupA{Luminance}, \GroupB{Contrast}, \GroupC{Chrominance}, \GroupD{Blur}, and \GroupE{Spatial Information}.}
%     \label{fig:feature-add}
%     \vspace{-3.5mm}
% \end{figure*}

\begin{figure*}
\centering
\includegraphics[width = \linewidth]{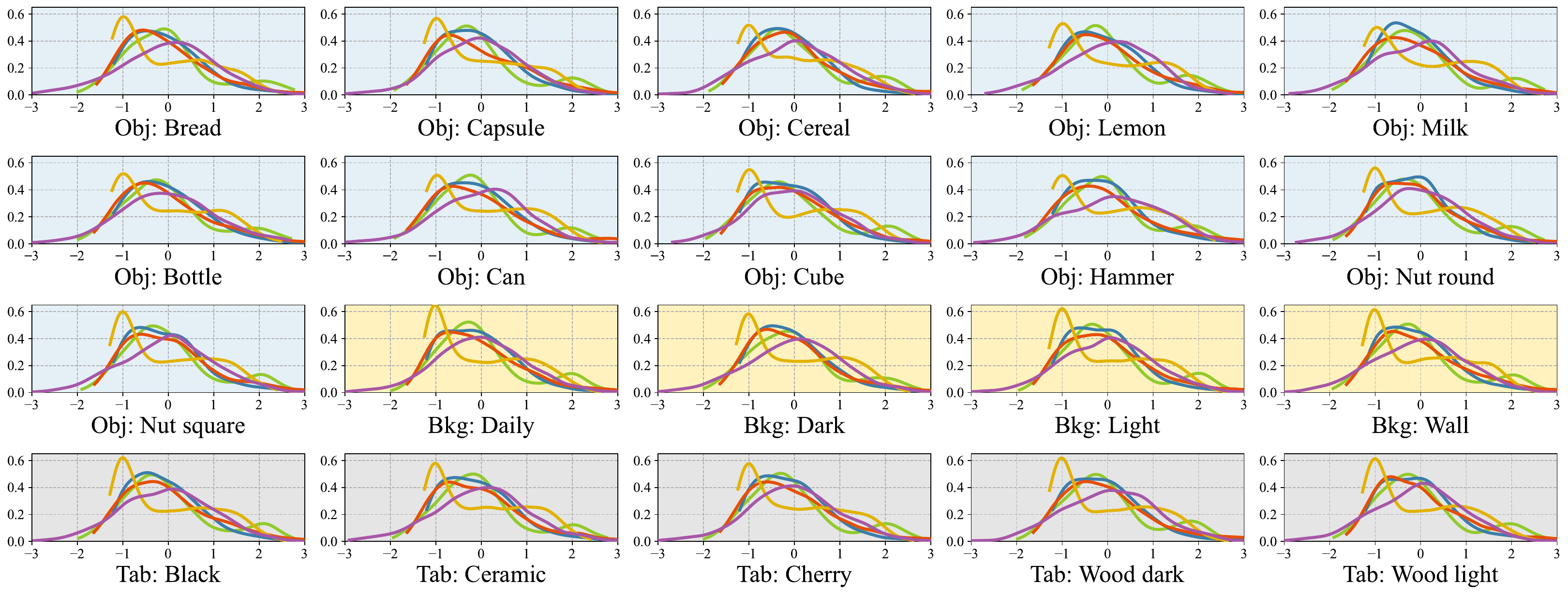}
\vspace{-3mm}
\caption{Low-level distributions for each Main Object/Table/Background. Results show the low-level feature differences are not due to the image content itself, but to compression inside the content. [Keys: \GroupA{Luminance}, \GroupB{Contrast}, \GroupC{Chrominance}, \GroupD{Blur}, \GroupE{Spatial Information}.]}
%\vspace{-1mm}
\label{fig:low}
\end{figure*}
% 结果显示不是图像本身的内容，而是对同一内容的压缩，造成了low-level的特征差异

\begin{figure*}
\centering
\includegraphics[width = \linewidth]{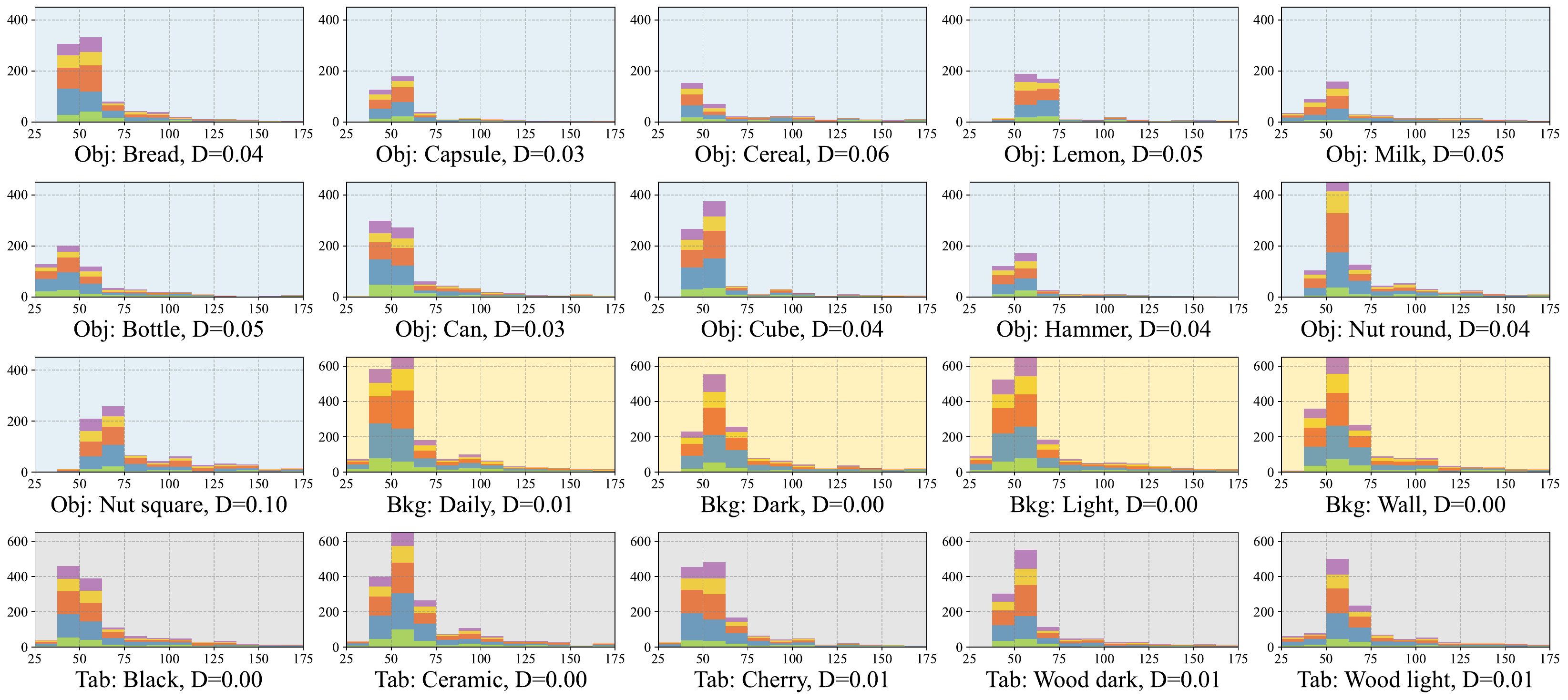}
\vspace{-3mm}
\caption{For successfully executed instances, the Step follows a similar distribution for the same Main Object/Table/Background, which is not correlated to bitrate according to Wasserstein Distance (D). [Keys: Below \GroupA{0.02}/\GroupB{0.04}/\GroupC{0.06}/\GroupD{0.08}/\GroupE{0.10} bpp range.]}
\vspace{-1mm}
\label{fig:step}
\end{figure*}

\section{Bitrate Analysis}

In the main text, we conduct a comprehensive analysis of bitrate. Based on the actual application scenarios of Embodied AI and current IoT communication protocols, we derived the bpp range that EmbodiedComp can operate on. This range requires far more extreme compression than current codec applications (e.g. broadcast TV and streaming media). Here, based on the formulas in the main text, we list several possible scenarios and their corresponding bpp:
\begin{itemize}
    \item Ideal: 10 IoT nodes blanket a two-storey house and enjoy a 30 dB SNR. (0.137bpp)
    \item Assisted-Living Flat: 15 health and safety gadgets form a 20 dB mesh. (0.061bpp)
    \item Smart Office: 35 desk-level sensors share 5 GHz Wi-Fi at 24 dB. (0.031bpp)
    \item Micro-Market: 40 shelf, fridge and camera tags pull 20 dB on sub-GHz. (0.023bpp)
    \item Vineyard Plot: 30 soil and weather probes reach the gateway at 15 dB. (0.023bpp)
    \item Extreme: a 50-device studio flat drops to 15 dB, while a 50-node open office. (0.014bpp)
\end{itemize}
The SNR sweep spans 15 dB—where IoT antennas barely lock—to 30 dB of clean indoor reception, while the device count ranges from 10 (below which the system collapses into point-to-point links) to the 50-node limit of the deployed mesh protocol. These bounds rarely push bitrate to the theoretical extreme of 0.014 bpp, yet the overwhelming majority of operational points fall below 0.06 bpp and often below 0.03 bpp—\textbf{exactly the interval where our rate–performance curves reveal the first kink and the subsequent vertical drop.} Embodied Image Compression is therefore not an academic contrivance; it is the prerequisite for bringing multi-agent Embodied AI out of the laboratory and into bandwidth-constrained, interference-prone Real-world networks.

\section{SR \& Steps Evaluation Analysis}

This Section provides an additional analysis of the properties of each main object, table, and background. First, their low-level feature distributions are shown in Figure \ref{fig:low}. Here, `Luminance' and `Contrast' follow their literal definitions; `Chrominance' indicates the strength of the color channel; `Blur' denotes the information density filtered by the Sobel operator; and `Spatial Information' stands for the texture diversity of the images. It can be seen that, across the entire EmbodiedComp dataset, only Luminance and Spatial Information exhibit slight variations among different objects/tables/backgrounds, while the remaining distributions are almost identical. This indicates differences in low-level features mainly stem from compression-induced distortions rather than from the reference image itself.

Furthermore, for samples that have been successfully executed, Figure \ref{fig:step} statistics the required Steps, with five colors representing different bitrate ranges. The similarity of step-count distributions across the five bitrates is measured by the Wasserstein Distance, ranging from 0 to 1. Horizontally comparing each sub-figure, different objects/tables/backgrounds show certain differences in step consumption, especially for objects. Within each sub-figure, however, different bitrates do not introduce noticeable changes in step count; the distance never exceeds 0.1.

Combining the two parts above, we conclude that compression primarily alters the low-level features of objects, thereby causing the performance degradation of VLA. This degradation is mainly reflected in SR rather than Step. Only because failed samples are penalized to the maximum of 250 steps does Step also increase; once the execution is correct, no additional Step consumption is incurred.

\section{Disclaimer}

The reported metrics are benchmark-specific in the NB-IoT protocol, and may not predict performance on other hardware, network conditions, or sensor configurations; they are intended solely for codec comparison and not as absolute capability statements. Human is only involved in annotation, and no experiment is conducted on animals, or safety-critical systems, and nothing herein should be construed as criticism of any VLA architecture.  By illuminating the precise bitrate cliff at which cloud-edge collaboration falters, we aim to inspire codecs that erase this cliff—so bandwidth-starved factories, hospitals, and homes can become everyday arenas where Embodied AI safely serves.

% 瞄准于ICM中genreal machine到Embodied AI的演进, 我们首次提出了Embodied Image Compression这一全新任务，旨在让codec更好的服务于Embodied AI。首先，我们建立了EmbodiedComp测试基准，根据RVS的特性，将codec放在闭环里评估；其次，我们论证了Embodied AI使用codec的必要性，并划定了其工作的码率区间。最后，我们验证了现有前沿算法的表现，实验证明上述区间存在一个关键拐点，使得现有VLA失效。我们衷心希望EmbodiedComp能启发更好的ICM方法，推广Embodied AI的应用。

{
    \small
    \bibliographystyle{ieeenat_fullname}
    \bibliography{main}
}


\end{document}